\documentclass{article}

\usepackage{arxiv}

\usepackage[utf8]{inputenc} 
\usepackage[T1]{fontenc}    
\usepackage{hyperref}       
\usepackage{url}            
\usepackage{booktabs}       
\usepackage{amsfonts}       
\usepackage{nicefrac}       
\usepackage{microtype}      
\usepackage{lipsum}
\usepackage{graphicx}
\graphicspath{ {./images/} }

\title{PP-GAN : Style Transfer from Korean Portraits to ID Photos Using Landmark Extractor with GAN}

\author{
 Jongwook Si \\
  Dept. Computer AI Convergence Engineering\\
  Kumoh National Institute of Technology\\
  Gumi, KOREA 39177 \\
  \texttt{jwsi425@kumoh.ac.kr} \\
   \And
 Sungyoung Kim \\
  Dept. Computer Engineering\\
  Kumoh National Institute of Technology\\
  Gumi, KOREA 39177 \\
  \texttt{sykim@kumoh.ac.kr} \\
  \And
}

\begin{document}
\maketitle
\begin{abstract}
The objective of a style transfer is to maintain the content of an image while transferring the style of another image. However, conventional research on style transfer has a significant limitation in preserving facial landmarks, such as the eyes, nose, and mouth, which are crucial for maintaining the identity of the image. In Korean portraits, the majority of individuals wear "Gat", a type of headdress exclusively worn by men. Owing to its distinct characteristics from the hair in ID photos, transferring the "Gat" is challenging. To address this issue, this study proposes a deep learning network that can perform style transfer, including the "Gat", while preserving the identity of the face. Unlike existing style transfer approaches, the proposed method aims to preserve texture, costume, and the "Gat" on the style image. The Generative Adversarial Network forms the backbone of the proposed network. The color, texture, and intensity were extracted differently based on the characteristics of each block and layer of the pre-trained VGG-16, and only the necessary elements during training were preserved using a facial landmark mask. The head area was presented using the eyebrow area to transfer the "Gat". Furthermore, the identity of the face was retained, and style correlation was considered based on the Gram matrix. The proposed approach demonstrated superior transfer and preservation performance compared to previous studies. 
\end{abstract}


\section{Introduction}
With the advent of modern technologies, such as photography, capturing the appearances of people has become effortless. However, when these technologies were not developed artists would paint portraits of individuals. Such a painting is called a portrait, and because of the invention of photography, modern portraits have become a new field of art. However, all famous figures from the past were handed down in pictures. The main purpose of paintings is to depict politically famous figures, but in modern times, the purpose has expanded to the general public. Although the characteristics of portraits by period and country are very different, most differ greatly from the actual appearance of the characters unless they are surrealistic works. Korean portraits differ considerably depending on time and region. Fig. 1(a) shows a representative work of portraits from the Goryeo Dynasty. This work is a portrait of Ahn Hyang, a Neo-Confucian scholar from the mid-Goryeo period. Fig. 1(b) is a portrait of the late Joseon Dynasty, which indicates that there is a large difference in the preservation conditions and drawing techniques. In particular, in Fig. 1(b), the "Gat" on the head is clearly visible.

\begin{figure}[htb!]
    \centering
    \includegraphics[width=10cm]{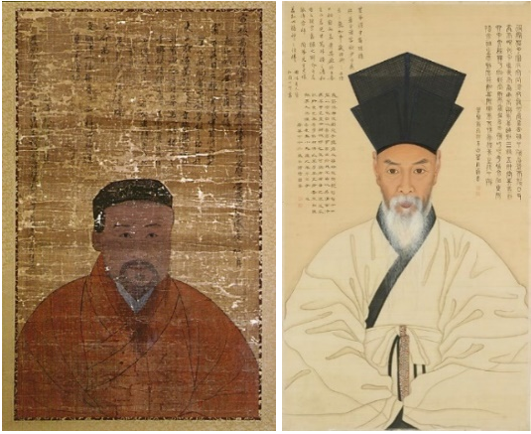}
    \caption{The left photo (a) is a portrait of Hyang An (1243 ~ 1306) in the mid-Goryeo dynasty and the right photo (b) is a portrait of Chae Lee (1411 ~ 1493) in the late Joseon Dynasty}
\end{figure}

Prior to the Three Kingdoms Period, Korean portrait records were absent, and only a limited quantity of portraits were preserved during the Goryeo Dynasty [1]. In contrast, the Joseon Dynasty produced numerous portraits with different types delineated according to their social status. Furthermore, works from the Joseon era exhibit a superior level of painting, in which facial features are rendered in a greater detail than in earlier periods.
A portrait exhibits slight variations in the physical appearance of a person, but it uniquely distinguishes individuals akin to a montage. Modern identification photographs serve a similar purpose and are used as identification cards, such as driver's licenses and resident registration cards. Old portraits may pique interest in how one appears in such artwork, for which style transfer technology can be used. Korean portraits may be used to provide the style of ID photos; however, the custom of wearing a “Gat” headgear renders transferring the style from Korean portraits to ID photos using previous techniques challenging. While earlier studies have employed global styles or partial styles to transfer onto content images, the distinct styles of texture, attire, and “Gat” must be considered simultaneously for Korean portraits. By independently extracting several styles from the style image, transferring the age, hairstyle, and costume of the person in a portrait onto the ID photo is possible. Fig. 2 depicts the outcome of style transfer using CycleGAN, a prevalent method for style transfer, revealing the difficulty of achieving adequate style transfer when numerous styles are involved. In this study, we introduce a method for high-quality style transfer of Korean portraits, which overcomes the limitations of previous research to accurately preserve facial landmarks and produce realistic results.

\begin{figure}[htb!]
    \centering
    \includegraphics[width=12cm]{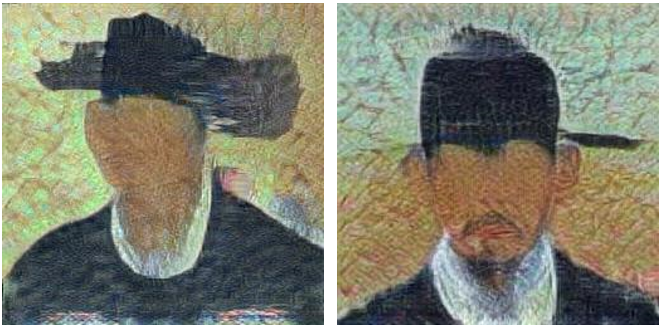}
    \caption{Results of style transfer from Korean portraits to ID photos using CycleGAN}
\end{figure}

Style transfer techniques, such as GAN, are commonly used based on facial datasets, but maintaining the identity of the person is crucial for achieving high-quality results. Existing face-based style transfer studies only consider facial components, such as eyes, nose, mouth, and hair, when transferring styles onto content images. In contrast, this study aims to transfer multiple styles, including Gats and costumes, simultaneously. To accomplish this, we propose an enhanced GAN-based network for style transfer that generates a mask using landmarks and defines a new loss function to perform style transfer based on facial data. we define the proposed method, "Style Transfer from Korean Portraits to ID Photos Using Landmark Extractor with GAN" as PP-GAN. The primary contribution of this study is the development of a novel approach to style transfer that considers multiple styles and maintains the identity of a person.

•	The possibility of independent and arbitrary style transfer to a network trained with a small dataset has been demonstrated. \\
•	This study is the first attempt at arbitrary style transfer in Korean portraits, which was achieved by introducing a new combination of loss functions.\\
•	The generated landmark mask improved the performance of identity preservation and outperformed previous methods [2].\\
•	New data on upper-body Korean portraits and ID photo were collected for this study.\\

\section{Related Works}
\label{sec:headings}
Research on style transfer can be categorized into two main groups: those based on convolutional neural Networks [4-6], [8-13] and those based on general adversarial networks [3], [14-23].

\subsection{CNN-Based Style Transfer}
\label{sec:headings}
AdaIN [4] suggested a method of transferring style at high speed using statistics in feature maps of content and style images. This is one of the earlier studies on style transfers. S. Huang et. al. [5] used the correlation between the content feature map and scaling information of the style feature map for the fusion of content and style. In addition, the order statistics method, called “Style Projection”, demonstrated the advantages and results of fast training speed. Zhu et. al. [6] maintained structural distortion and content by presenting a style transfer network that could preserve details. In addition, by presenting the refined network, which modified the VGG-16[7], the style pattern was preserved via spatial matching of hierarchical structures. Elad et. al. [8] proposed a new style transfer algorithm that expanded the texture synthesis work. It aimed to create images of similar quality and emphasized a consistent way of creating rich styles while keeping the content intact in the selected area. In addition, it was fast and flexible to process any pair of content and style images. S. Li et. al. [9] suggested a style transfer method for low-level features to express content images in a CNN. Low-level features dominate the detailed structure of new images. A Laplacian matrix was used to detect edges and contours. It shows a better stylized image, which can preserve the details of the content image and remove artifacts. Chen et. al. [10] proposed a stepwise method based on a deep neural network for synthesizing facial sketches. It showed better performance by proposing a pyramid column feature to enrich the parts by adding texture and shading. Fast Art-CNN [11] is a structure for fast style transfer performance in the feedforward mode while minimizing deterioration in image quality. It can be used in real-time environments as a method for training deconvolutional neural networks to apply a specific style to content images. X. Lio et. al. [12] proposed an architecture that includes geometric elements in the style transfer. This new architecture can transfer textures into distorted images. In addition, because the content/texture-style/geometry style can be selected to be entered in triple, it provides much greater versatility to the output. P. Kaur et. al. [13] proposed a framework that solves the problem of realistically transferring the texture of the face from the style image to the content image without changing the identity of the original content image. Changes around the landmark are gently suppressed to preserve the facial structure so that it can be transferred without changing the identity of the face. 

\subsection{GAN-Based Style Transfer}
\label{sec:headings}
APDrawingGAN [14] improved the performance by combining global and regional networks. High-quality results were generated by measuring the similarity between the distance transform and artist drawing. Zheng Xu et. al. [15] used a generator and discriminator as conditional networks. Subsequently, the mask module for style adjustment and AdaIN [4] for style transfer performed better than existing GAN. S3-GAN [16] introduced a style separation method in the latent vector space to separate style and content. A style-transferred vector space was created using a combination of separated latent vectors. CycleGAN [3] proposes a method for converting a style to an image without a pair of domains. While training the generator mapping to X → Y, reverse mapping to Y → X is performed. In addition, the cycle consistency loss was designed such that an input image and its reconstructed image could be identical when the transferred style was removed through reverse mapping. In SLGAN [17], a style-invariant decoder was created by a generator to preserve the identity of the content image and introduce a new perceptual makeup loss, resulting in high-quality conversion. Some attempts have been made to maintain facial landmarks in style transfer studies aimed at makeup [18-20] or aging [21-22]. BeautyGAN [18] defines instance and perceptual loss to change the makeup style while maintaining the identity of the face, thereby generating high-quality images and maintaining the identity. Paired-CycleGAN [19] trains two generators simultaneously to convert the makeup styles of other people from portrait photos. Stage 1 was used as a pair of powers through image analogy, and as an input of Stage 2, it showed excellent results by calculating identity preservation and style consistency compared to the power of Stage 1. Landmark-CycleGAN [20] showed incorrect results owing to the distortion of the geometrical structure while converting a face image to a cartoon image. To solve this problem, local discriminators have been proposed using landmarks to improve performance. Palsson's et. al. [21] suggested Group-GAN, which consisted of several models of CycleGAN[3], to integrate pre-trained age prediction models and solve the face aging problem. Wang et. al. [22] proposed a method for interconverting edge maps to a CycleGAN-based E2E-CycleGAN network for aging. The old face was generated using the identity feature map and result of converting the edge map using the E2F-pixelHD network. Yi et. al. [23] proposed a new asymmetric cycle mapping that forced the reconstruction information to be shown and included only in optional facial areas. Portrait images generated along with a localized discriminator for landmark and style classifiers were introduced. Considering the style vector, portraits were generated in several styles using a single network. They attempted to transfer the style of the portrait similar, which is similar to the purpose of our study. However, in this study, not only the portrait painting style but also the Gat and costume are transferred together.

\section{Background}
\subsection{VGG-16}
\label{sec:headings}
The VGG-16 [7] network is a prominent computer vision model that attained a 92.7\% Top-5 accuracy in the ImageNet Challenge competition by receiving an RGB image with dimensions of 224 × 224 as input, containing 16 layers in a configuration of 13 convolution layers and three FC layers. The convolution filter measures 3 × 3 pixels and maintains fixed strides and padding at 1. The activation function employed in the network is ReLU, and the pooling layer is max pooling, which is set to a fixed stride of 2 on 2 × 2. The closer it is to the input layer, the more low-level information the feature map contains, such as color and texture. of the image, and the closer it is to the output layer, thus providing higher-level information, such as shape. The pre-trained VGG-16 [7] is used in this study to preserve facial and upper-body content and transfer the style efficiently.

\subsection{Gram matrix}
\label{sec:headings}
The Gram matrix is a valuable tool for representing the color distribution of an image. This enabled the computation of the overall color and texture correlation between the two images. Leon et. al.[24] demonstrated that the style transfer performance can be improved using a Gram matrix on feature maps from various layers. Fig. 3 illustrates the process of calculating the Gram matrix, which involves converting each channel of a color image into a 1D vector, followed by obtaining the matrix by multiplying the H×W matrix with its transpose. The Gram matrix is a square matrix with channel size as its dimension. As the corresponding values in the Gram matrix of the two images become more similar, the color distribution of the images also becomes more similar.

\begin{figure}[htb!]
    \centering
    \includegraphics[width=13cm]{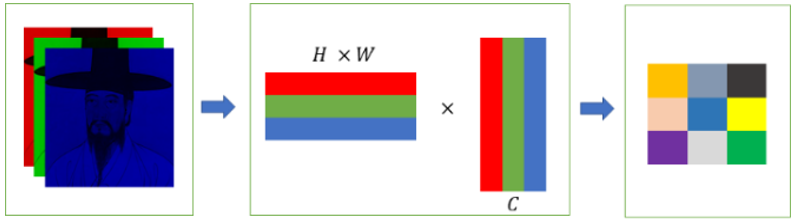}
    \caption{The process of calculating into a gram matrix for Korean portraits}
\end{figure}

\subsection{Face Landmark}
\label{sec:headings}
Facial landmarks, such as the eyes, nose, and mouth, play a vital role in identifying and analyzing facial structures. To detect the landmarks, this study employed the Shape Predictor of 68 face landmarks [25], which generated 68 x and y coordinates of the crucial facial components, including the jaw, eye, eyebrow, nose, and mouth and also provided the locations of the face. Subsequently, the coordinates obtained from the predictor were used to create masks for the eyes, nose, and mouth, as shown in Fig. 4.

\begin{figure}[htb!]
    \centering
    \includegraphics[width=10cm]{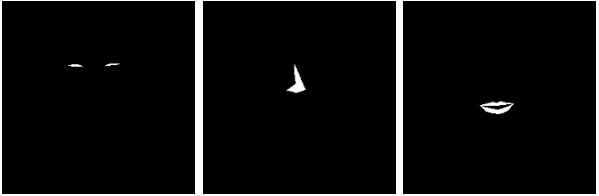}
    \caption{Masks for the eyes, nose, and mouth created using the coordinates returned by Shape Predictor 68 Face Landmarks}
\end{figure}

\subsection{Image Sharpening}
\label{sec:headings}
Image sharpening is considered a high-frequency emphasis filtering technique, which is employed to enhance image details. High frequency is characterized by the changes in brightness or color occurring locally, and it is useful in identifying facial landmarks. Image sharpening can be achieved using high-boost filtering. It involves generating a high-pass image by subtracting a low-pass image from an input image as shown in Eq. (1). A high-frequency emphasized image is obtained by multiplying the input image with a constant during this process.

\begin{equation}
g(x,y) = Af(x,y) - f_L(x,y) 
\end{equation}

Mean filtering is a low-pass filtering technique, and the coefficients of the filter can be determined using Eq. (2). The sharpening strength of the input image is controlled by the value of $\alpha$, where $9A-1$ is set to $\alpha$. A high $\alpha$ value results in a decrease in the sharpness level owing to the high ratio of the original image to the output. Conversely, a small $\alpha$ value results in a reduction in contrast, owing to the removal of numerous low-frequency components. 
To have similar structures between portrait images and ID photos, portrait images are cropped around the faces, as the face occupies a relatively small area. In contrast, ID photos are resized so that they have the same size, both horizontally and vertically, instead of being cropped. However, this resizing can make extracting facial landmarks difficult. Therefore, image sharpening is performed in the present study. This process is necessary to ensure that facial landmarks are extracted well from ID photos, as shown in Fig. 5, where the difference in facial landmark extraction with and without image sharpening is illustrated.

\begin{equation}
    A
    \left[
    \begin{array}{ccc}
        0 & 0 & 0 \\
        0 & 1 & 0 \\
        0 & 0 & 0 \\
    \end{array}
    \right]
    - 1/9
    \left[
    \begin{array}{ccc}
        1 & 1 & 1 \\
        1 & 1 & 1 \\
        1 & 1 & 1 \\
    \end{array}
    \right]
   = 1/9
    \left[
    \begin{array}{ccc}
        -1 & -1 & -1 \\
        -1 & 9A-1 & -1 \\
        -1 & -1 & -1 \\
    \end{array}
    \right]
   \rightarrow
   \left[
    \begin{array}{ccc}
        -1 & -1 & -1 \\
        -1 & \alpha & -1 \\
        -1 & -1 & -1 \\
    \end{array}
    \right]
\end{equation}

\begin{figure}[htb!]
    \centering
    \includegraphics[width=13cm]{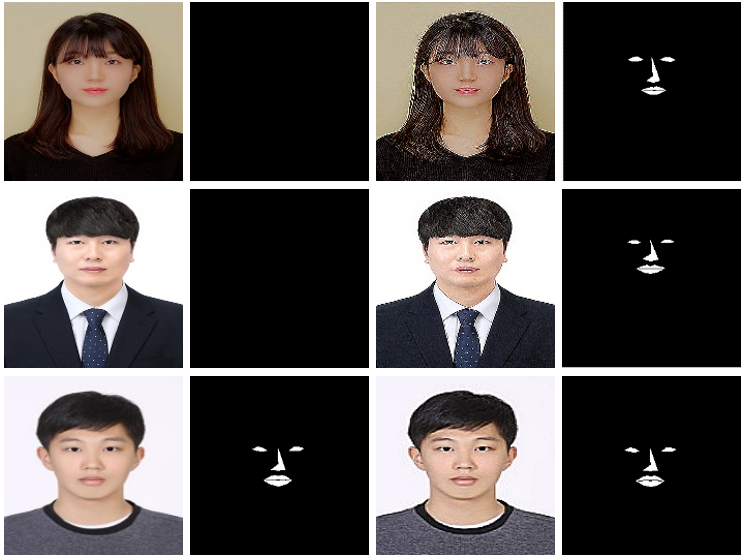}
    \caption{Result of landmark mask generation according to the use of high boost filtering (The first and third columns are the original and the high boost filtered image, respectively, and the second and fourth columns show the masks with the detected landmark for each corresponding image)}
\end{figure}

\section{Proposed Method}
\label{sec:headings}

\subsection{Network}
\label{sec:headings}

The primary objective of the proposed method is to achieve a style transfer of ID photos to Korean portraits. Let $X$ and $Y$ indicate the domains of the three-dimensional color image and the Korean portrait, respectively. These domains are subsets of $X \subseteq \mathbb{R}^{H \times W \times C}$ and $Y \subseteq \mathbb{R}^{H \times W \times C}$ and have a set relationship such that $x \in X$ and $y \in Y$.

The CycleGAN [3] network is limited in performing style transfer owing to its training over the entire domain. Therefore, the proposed method adopts a Dual I/O generator from BeautyGAN [18], which has a stable discriminator that enables mapping training between two domains and style transfer. Additionally, the proposed method incorporates VGG-16 ,a gram matrix, and a landmark extractor to improve performance. Fig. 6 depicts the overall structure of the proposed method.

\subsubsection{Generator}

The generator is trained to perform $(X, Y) \rightarrow (Y, X)$ mapping, resulting in a fake image $G(x, y) = (x_y, y_x)$ with content $X$ and style $Y$, which is evaluated in this study. Contents $Y$ and $X$ are used to generate another fake image $y_x$. This study focuses only on $x_y$ results, even though the network structure can generate results in both directions.
The image recovered by the Dual I/O generator performing style transfer and the input image must be identical. With an input size of $(256, 256, 32)$, $x$ and $y$ pass through three convolution layers each, resulting in a size of $(64, 64, 128)$. The $x$ and $y$ results are concatenated to produce a size of $(64, 64, 256)$, which is restored to the original size through the deconvolution layer, allowing style transfer through nine residual blocks. This result represents a fake image of a style transferred and represents the result of the proposed method.
Therefore, the generator deceives the discriminator by generating fake images that appear real, resulting in more natural and higher-performance results.

\subsubsection{Discriminator}

The network structure includes two discriminators that are trained to classify the styles of fake and real images generated by the generator. The discriminator consists of five convolution layers and aims to distinguish styles. The input image size is $(256, 256, 3)$, and the network result size is $(30, 30, 1)$. The first four convolution layers, excluding the last layer, perform Spectral Normalization [26] to improve performance and maintain a stable distribution of the discriminator in a high-dimensional space.
The discriminator is defined as follows: $D_x$ classifies $x_y$ as fake and $y$ as real, whereas $D_y$ classifies $y_x$ as fake and $x$ as real. Finally, the PatchGAN [27] is applied to produce the discriminator output, which is the final judgment result of the discriminator on the input image.

\begin{figure}[htb!]
    \centering
    \includegraphics[width=16cm]{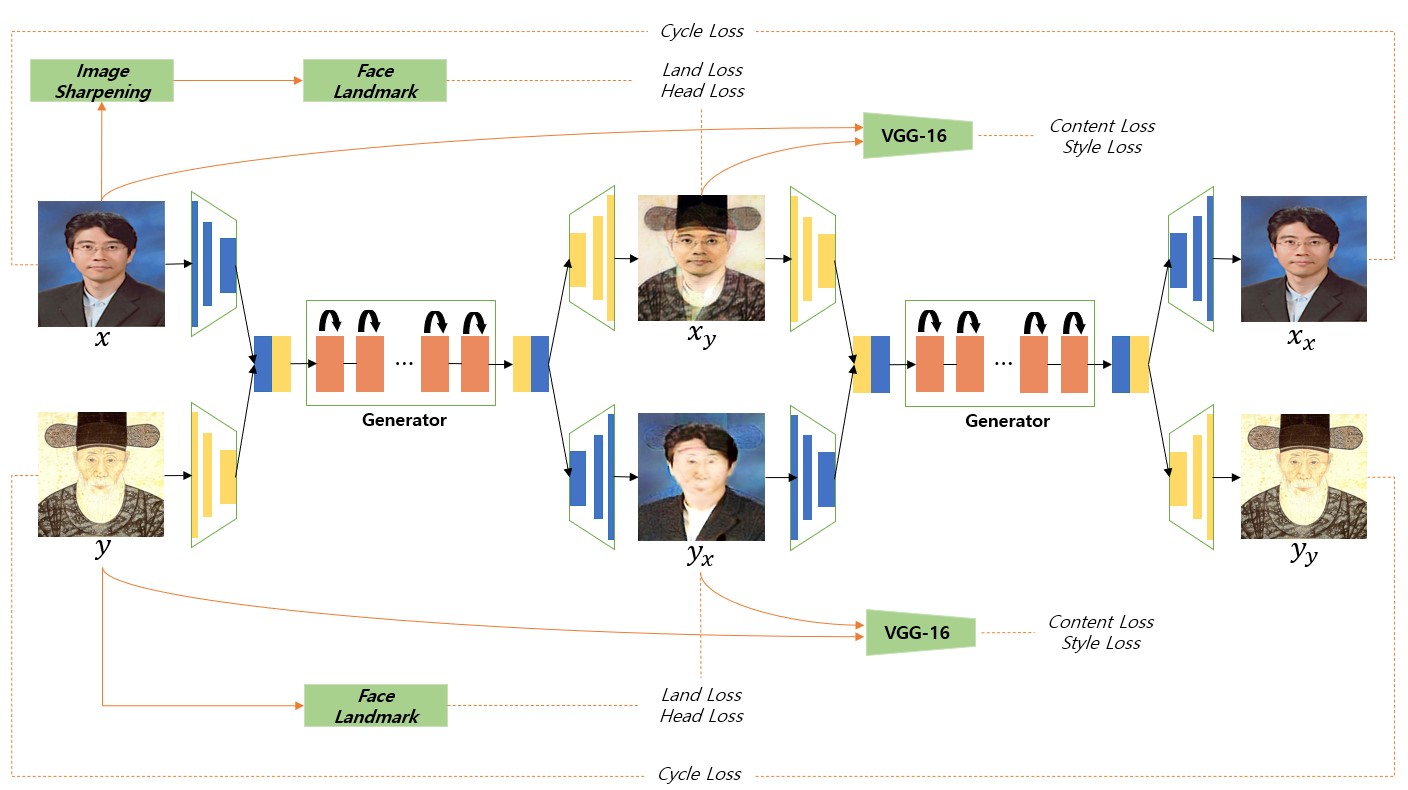}
    \caption{Overall structure of the system proposed in this study}
\end{figure}

\subsection{Loss Functions}
\label{sec:headings}
In this study, we propose a loss function for transferring ID photos to arbitrary Korean portrait styles. Six loss functions, including the loss function of the new approach, are used to generate good results.
CycleGAN introduced the concept of setting the result as the input of the generator again through a cycle structure, which should theoretically produce the same output as the original image. Therefore, in this study, we define the recovered result as the cycle loss, which consists of a loss function designed to reduce the difference between the input and output images. In particular, it can be expressed as $x \approx G(G(x,y)) = G(x_y, y_x) = x_x$ and $y \approx G(G(y,x)) = G(y_x, x_y) = y_y$. This can be expressed as Eq. (3).\\

\begin{equation}
L_{cy} = \mathbb{E}_{x \sim P(X)} \|x_x - x\| + \mathbb{E}_{y \sim P(Y)} \|y_y - y\}
\end{equation}

The existing style transfer method distorts the shape of the face geometrically, leading to difficulties in recognizing the face shape. To maintain the identity of the character, a new condition is required. Hence, this study defines land loss based on a face landmark mask, which helps in preserving the eyes, nose, and mouth while enhancing the performance of style transfer. Land loss is defined by the mathematical expression Eq. (4) in this study.

\begin{equation}
L_l = L_{l_{eye}} + L_{l_{nose}} + L_{l_{lip}}
\end{equation}

Land loss is a function that aims to maintain the landmark features of the input and output images generated by the generator. The pairs of images $(x_y, x)$ and $(y_x, y)$ contain the same content with different styles, and the landmark shapes are identical. The masks $M_{fX}$ and $M_{fY}$, generated for the eye, nose, and mouth areas, are used to calculate the area, as discussed in Section 3. Using a pixel-wise operation, each area of the eye, nose, and mouth is processed using a face landmark mask, and a loss function is defined to minimize the difference in pixel values. This process is expressed in Eq. (5). The difference for each landmark is based on L1 Loss.

\begin{equation}
L_f = \mathbb{E}_{x \sim P(X)} \|x_y \odot M_{fX} - x \odot M_{fX}\|_1 + \mathbb{E}_{y \sim P(Y)} \|y_x \odot M_{fY} - y \odot M_{fY}\|_1, \quad f = \{l_{eye}, l_{nose}, l_{lip}\}
\end{equation}

The method proposed in this study differs greatly from previous style transfer research, which requires some content of the style target image rather than ignoring it and only considering the color relationship. In particular, for Korean portraits, the style of the Gat and clothes must be considered in addition to image quality, background, and overall color. However, the form of the Gat varies widely, which is difficult to detect due to the differences in wearing position, while the hair of Korean portraits and ID photos have completely different shapes. To address this, a head loss is proposed to minimize the difference between the head area of the result and style images, with the head area divided into the Gat and hair areas, represented by masks $M_{ht}$ and $M_{hr}$. Head loss uses the fact that the Gat does not cover the eyebrows; therefore, the feature point located at the top of the coordinates corresponding to the eyebrows is used to define the head area, which is then used to transfer the corresponding style to the resulting image. This is expressed in Eq. (6).

\begin{equation}
L_h = \mathbb{E}_{x \sim P(X)} \|x_y \odot M_{ht} - y \odot M_{ht}\|_1 + \mathbb{E}_{y \sim P(Y)} \|y_x \odot M_{hr} - x \odot M_{hr}\|_1
\end{equation}

To preserve the overall shape of the character and enhance the performance of style transfer, content loss and style loss are defined using a specific layer of VGG-16 in this study. The pre-trained network contains low- and high-level information, such as colors and shapes, which appear differently depending on the layer location. Low-level information is related to style, and high-level information is related to content. Conversely, high-level layers represent the image characteristics. Therefore, the content and style losses are configured based on the layer characteristics. Style loss is defined using a gram matrix, which is obtained by computing the inner product of the feature maps. The best set of layers obtained through the experiment is used to define style loss, as shown in Eq. (7), where $N$ and $M$ represent the product and channel of each layer, respectively, and $g$ represents the gram matrix of the feature map. By training to minimize the difference in the gram matrix between the feature maps for both sides ($x_y$ and $y_x$), the style of $y$ can be transferred to $x$.

\begin{equation}
L_s = \frac{1}{4N^2M^2} \sum \left[ \left(g_i(x_y) - g_i(y)\right)^2 + \left(g_i(y_x) - g_i(x)\right)^2 \right]
\end{equation}

Content loss is defined as a method to minimize linear differences in feature maps at the pixel level. Because the style transfer aims to maintain the content of an image while transferring the style, it is not necessary to consider correlations. The equation for content loss is the same as that in Eq. (8). This is a critical factor in preserving the identity of a person; however, if the weight of this loss is extremely large, it can result in poor style transfer results. Therefore, appropriate hyperparameters must be selected to achieve the desired outcome.

\begin{equation}
L_c = \mathbb{E}_{x \sim P(X)} [l_i(x_y) - l_i(x)]^2 + \mathbb{E}_{y \sim P(Y)} [l_i(y_x) - l_i(y)]^2
\end{equation}

The generator loss is composed of cycle, land, head, style, and content losses, as expressed in Eq. (9). Each loss is multiplied by a different hyperparameter, and the sum of the resulting values is used as the loss function of the generator.

\begin{equation}
L_G = \lambda_{cy} L_{cy} + \lambda_{l} L_{l} + \lambda_{h} L_{h} + \lambda_{s} L_{s} + \lambda_{c} L_{c} 
\end{equation}

The discriminator loss solely comprises adversarial loss, which follows the GAN structure. The output of the discriminator is a $32 \times 32 \times 1$ result that is evaluated based on PatchGAN [27] to identify whether they are authentic or fake, considering every image PatchGAN [27]. The loss function used to train the discriminator is given by Eq. (10). The loss function is reduced if the patches of $x_y$ and $y_x$ are fake, and the patches of $x$ and $y$ are genuinely classified.

\begin{equation}
L_D = \mathbb{E}_{x \sim P(X)} [(D_x(y) - 1)^2 + (D_x(x_y))^2] + \mathbb{E}_{y \sim P(Y)} [(D_y(x) - 1)^2 + (D_y(y_x))^2]
\end{equation}

The total loss employed in this study is expressed by Eq. (11) and is composed of the generator and discriminator losses. The generator seeks to minimize the generator loss to generate style transfer outcomes, whereas the discriminator aims to minimize the discriminator loss to enhance its discriminative capability. A trade-off between the generator and discriminator performances is observed, where if one is improved, the other is diminished. Consequently, the total loss is optimized by forming a competitive relationship between the generator and discriminator, which leads to superior outcomes.

\begin{equation}
L_{Total} = \min_G \min_D (L_G + L_D)
\end{equation}

\subsection{Training}
The experimental environment in this study was conducted on a multi-GPU system using the GeForce RTX 3090 and Ubuntu 18.04 LTS operating system. As TensorFlow 1.x has a minimum version requirement for CUDA, the experiments were carried out using Nvidia-Tensorflow version 1.15.4.
Datasets of ID photos and Korean portraits were collected through web crawling using Google and Bing search engines. To improve the training performance, preprocessing was conducted to separate the face area from the whole body of the Korean portraits, which typically feature the entire body. Data augmentation techniques, such as left and right inversion, blur, and noise, were applied to increase the limited number of datasets. Gat preprocessing was also performed, as shown in Fig. 7, to facilitate the feature mapping.
Tab. 1 shows the resulting dataset consisting of 1,054 ID photos and 1,736 Korean portraits divided into 96\% training and 4\% test sets. Owing to the limited number of portraits, a higher ratio of training data was used, and no data augmentation was applied to the test set. As the number of combinations that could be generated from the test data was substantial ($X_{Test}$ × $Y_{Test}$), the evaluation was not problematic.
Previous research has emphasized the importance of data preprocessing, and the results of this study further support its impact on training performance.

\begin{figure}[htb!]
    \centering
    \includegraphics[width=9cm]{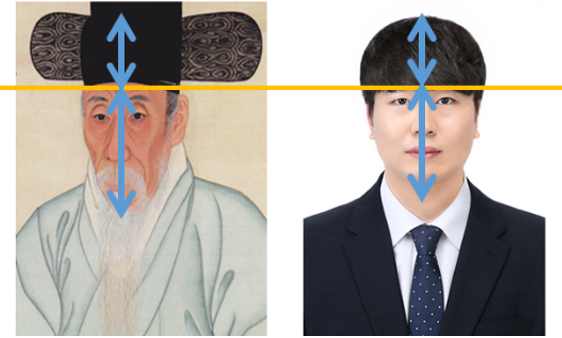}
    \caption{Examples of datasets preprocessing}
\end{figure}

\begin{table}[htbp]
 \caption{Detailed Datasets}
  \centering
  \begin{tabular}{cccc}
    \toprule
    Data     & Train     & Test & Sum \\
     \midrule
    ID Photos($X$) & 988 & 66 & 1,024\\

    Korean Portraits($Y$) & 1,714 & 22 & 1,736\\
     \midrule
     Sum($X+Y$) & 2,702 & 88 & 2,790\\
     \midrule   
     Combination($XY$) & 1,693,432 & 1,452 & -\\
    \bottomrule
  \end{tabular}
  \label{tab:table}
\end{table}

The proposed network was trained for 200 epochs using the Adam Optimizer. The initial learning rate was set to 0.0001 and linearly reduced to zero after 50\% of the training epoch for stable learning. To match the equality between the loss functions, \(\lambda_{cy}\) was set to 50, which resulted in a relatively lower value than the other losses. To increase the effect of style transfer, \(\lambda_s\) was set to 1 and \(\lambda_h\) was set to 0.5, which helped concentrate on the head area between the style transfers. Finally, the training proceeded by setting \(\lambda_c = 0.1\) and \(\lambda_l = 0.2\). The entire training process took approximately 6.5 hours. The results are presented in Fig. 8, which visually confirms that the proposed method shows a greater performance improvement than previous research [2]. While previous methods have focused only on style transfer, this study successfully maintained the identity of a person while transferring the style. The results show excellent outcomes in which the style is transferred while preserving the shape of the character in the content image. Additionally, the identity of the personnel is preserved, and the Gat is transferred naturally.

\begin{figure}[htb!]
    \centering
    \includegraphics[width=16cm]{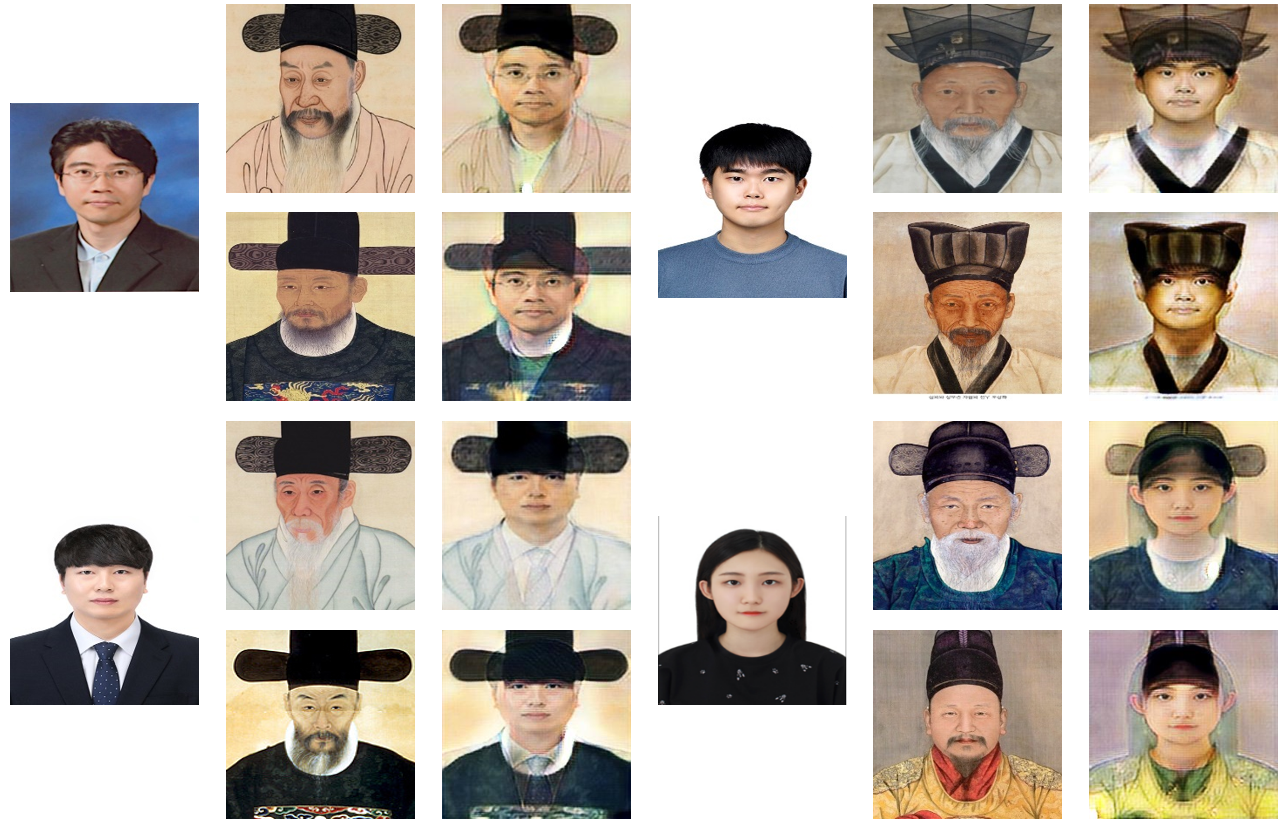}
    \caption{The result of the method proposed in this paper}
\end{figure}

\section{Experiments}
\label{sec:headings}

\subsection{Feature Map}
\label{sec:headings}
To perform style loss, this study adopted Conv2\_2 and Conv3\_2 from the VGG-16 layer, whereas Conv4\_1 was used for content loss. Although the front convolution layers contain low-level information, they are sensitive to change and difficult to train because of the large differences in pattern and color between the style and content images. To overcome this problem, this study utilizes a feature map located in the middle to extract low-level information for style transfer. The results of using a feature map not adopted for style loss are presented in Fig. 9 with a smoothing set to 0.8 on the tensorboard. Loss graphs were output for only ten epochs because of the failure of training when layers not used for Style Loss were utilized in the experiment.

•	The Conv2\_1 layer exhibits a large loss value and unstable behavior during training, indicating that training may not be effective for this layer.\\
•	Conv1\_2 is close to zero in most of the losses, but it cannot be said that the training proceeds because the maximum minimum differs by more than $10^4$ times owing to the very large loss of some data.\\
•	Conv1\_1 exhibits a high loss deviation and instability during training, similar to Conv2\_1 and Conv1\_2. Moreover, owing to its sensitivity to color, this layer presents challenges for training.

If Conv4\_1 is utilized as the style loss layer, it can transfer the style of the image content. However, because the feature map scarcely includes style-related content, the generator may produce images lacking style. Nevertheless, it is observed that transferring the style of the background is feasible as it corresponds to the overall style of the image and can be recognized as content because clothing style is not a prominent feature. Therefore, high-level layers, such as Conv4\_1, contain only the style of the background in the character content. The result of utilizing the feature map employed in the content loss for style loss is shown in Fig. 10. In general, most contents of the content image are preserved, whereas the style is marginally transferred. Hence, we proceed with using trainable layers, which results in stable training and enables us to transfer styles while conserving content.

\begin{figure}[htb!]
    \centering
    \includegraphics[width=15cm]{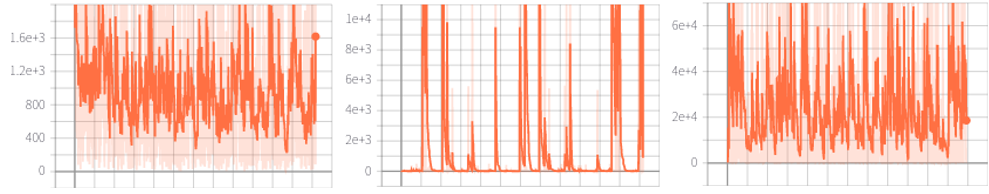}
    \caption{Result of a specific feature map experiment of VGG-16 for style loss}
\end{figure}

\begin{figure}[htb!]
    \centering
    \includegraphics[width=10cm]{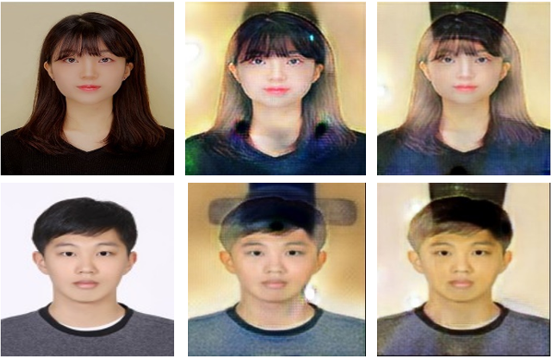}
    \caption{Results when the feature map used for Style Loss is used with the same layer as Content Loss (Column 1: Input Image; Column 2: and Column 3: output image using Conv4\_1)}
\end{figure}

\subsection{Ablation Study}
\label{sec:headings}
An ablation study was conducted on four loss functions, excluding Cycle Loss, to demonstrate the effectiveness of the loss function proposed in this study for Generator Loss. The results are presented in Fig. 11, where one row represents the use of the same content and style images. If $L_c$ is excluded, the character shape is not preserved, leading to poor results owing to the style concentration during training. Consequently, only facial components are transferred based on $L_l$. When $L_c$ and $L_l$ are excluded simultaneously, the style transfer outcome lacks facial components. Similarly, when $L_s$ is excluded, the result of the style transfer is of poor quality, with the character remaining almost the same. The use of $L_{cy}$ allows for style transfer of the background without the need for a separate loss function for style; however, owing to the main focus of training on the character shape, style transfer barely occurs, resulting in the creation of the Gat when $L_h$ is used. Excluding $L_l$ leads to unclear and blurred liver facial components, and the face color becomes bright. Therefore, $L_l$ plays a crucial role in preserving the character identity by making the liver face components more apparent. If $L_h$ is excluded, the head area becomes blurred or is not created, leading to unsatisfactory style transfer results. Unlike the overall style transfer, the Gat must be newly generated; thus, $L_s$ serves a different purpose. Therefore, the head area must be set separately, and $L_s$ can be used with $L_h$ to achieve this purpose. Consequently, the use of all the loss functions proposed in this study results in the best performance, generating natural images without bias in any direction.

\subsection{Performance Evaluation}
\label{sec:headings}
This study conducts a performance comparison with previous research [2] with the same subject, as well as an ablation study. Although there are diverse existing studies on style transfer, the subject of this paper differs significantly from them, leading to a comparison only with a single previous study [2].
Based on CycleGAN [3], evaluation in pairs could not be performed because of the impossibility of arbitrary style transfer. Thus, an evaluation survey was conducted with 59 students of different grades from the Department of Computer Engineering at Kumoh National Institute of Technology to evaluate the performance of the proposed method in terms of three items: the transfer of style ($S_{st}$), preservation of content ($S_{cn}$), and generation of natural images ($S_{nt}$). The survey was conducted online for 10 days using Google Forms, and the surveyors received and evaluated a combination of 10 results from a previous study [2] and 10 results from the proposed method. The survey results presented in Tab. 2 show that the proposed method outperforms the previous method [2] in all three aspects, with the highest difference in scores for preserving character content. In contrast, the previous method [2] failed to preserve the shape of the character during style transfer, leading to blurring or disappearance of the landmark of the face, which was not natural. However, the proposed method successfully preserved the character content and effectively transferred the style, producing relatively natural results. Thus, the proposed method showed better overall performance than the previous method [2].

\begin{table}[htbp]
 \caption{Survey Results}
  \centering
  \begin{tabular}{cccc}
    \toprule
    Score     & Previous method [2]    & Ours & Gap \\
     \midrule
    $S_{cn}$ & 2.510 & \textbf{3.867} & \underline{1.357}\\
    $S_{st}$ & 3.378 & \textbf{3.402} & 0.025\\
    $S_{nt}$ & 2.295 & \textbf{3.262} & 0.337\\
     \midrule
      Sum & 8.813 & \textbf{10.531} & 1.718\\
    \bottomrule
  \end{tabular}
  \label{tab:table}
\end{table}

\begin{figure}[htb!]
    \centering
    \includegraphics[width=15cm]{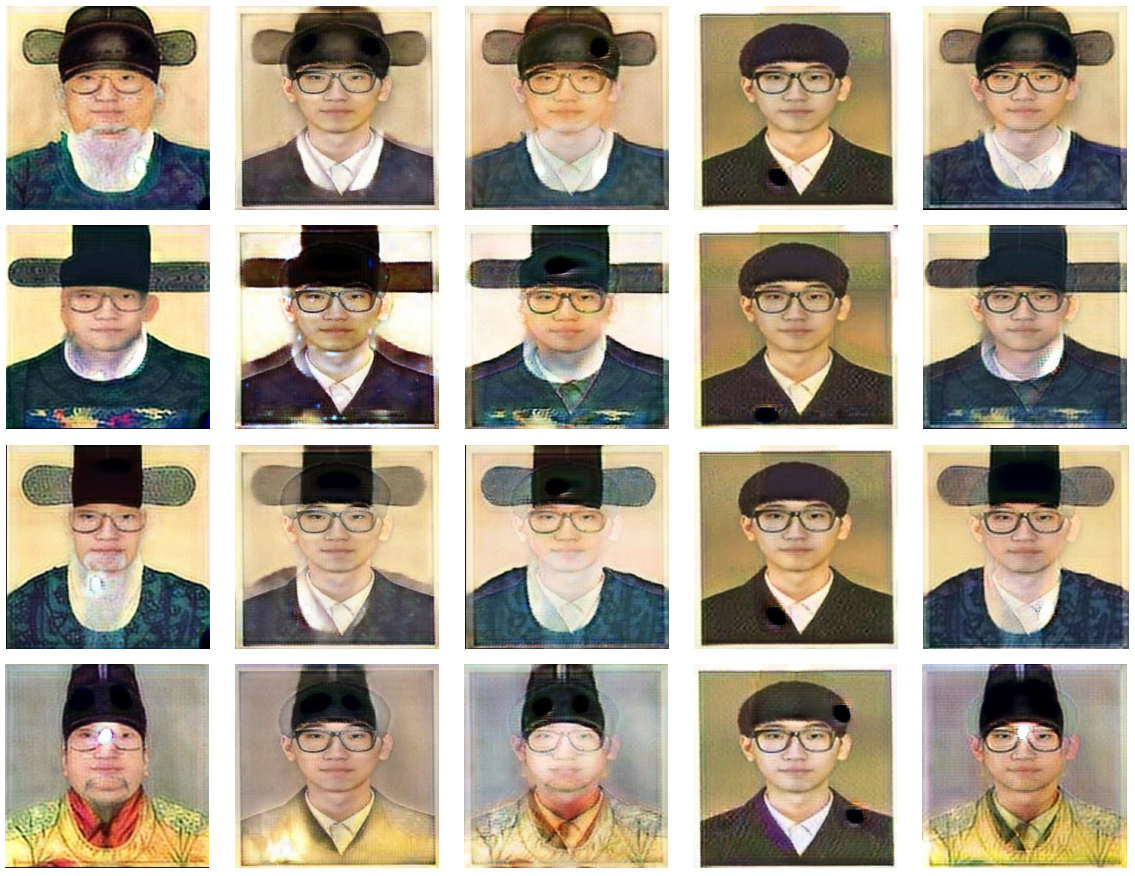}
    \caption{Results of the importance of the various loss functions that make up the Total Loss proposed in this paper. Column 1 to Column 4 Excluding the loss functions $L_c$, $L_s$, $L_l$, and $L_h$; Column 5 results generated using all loss functions}
\end{figure}

Peak signal-to-noise ratio (PSNR) and structural similarity index measure (SSIM) are commonly employed to measure performance, as observed in various studies [28-29]. However, when conducting style transfer while preserving content, it is crucial to ensure a natural outcome without any significant bias towards either the content or style. Consequently, to compare the performance, we propose new performance indicators that utilize a weighted arithmetic mean, which is further weighted by the median values of the PSNR and SSIM. The final result is obtained by combining the results of both content and style using the proposed performance indicator.
The PSNR is commonly used to assess the quality of images after compression by measuring the ratio of noise to the maximum value, which can be calculated using Eq. (12). The logarithmic denominator in this equation represents the average sum of squares between the original and compressed images, and a lower value indicates a higher PSNR and better preservation of the original image. By contrast, the SSIM is used to evaluate distortions in image similarity between a pair of images by comparing their structural, luminance, and contrast features. Eq. (13) is used to calculate the SSIM, which involves various probability-related definitions such as mean, standard deviation, and covariance.

\begin{equation}
    PSNR(A,B) = 10\log_{10}\left(\frac{MAX^2}{\sum{(A-B)^2}}\right) 
\end{equation}

\begin{equation}
    SSIM(A,B) = \frac{(2\mu_A\mu_B + C_1)(2\sigma_{AB} + C_2)}{(2\mu_A^2 + \mu_B^2 + C_1)(\sigma_A^2 + \sigma_B^2 + C_2)} 
\end{equation}

To evaluate the performance, the indicators are sorted in ascending order, resulting in a sequence of values denoted as $[x_1,x_2,x_3,x_4,x_5]$, where $x_3$ represents the best result. To assign more weight to the median value, a weight vector ($w$) of $[10, 25, 50, 25, 10]$ is assigned, and the weighted arithmetic mean is calculated using Eq. (14). The performance is evaluated using Eq. (15), which calculates the square of the difference between the weighted arithmetic mean and PSNR and SSIM values. The resulting value indicates the degree of performance, with smaller values indicating better performance. The difference from the average weight ($w_{avg}$) is squared and added, resulting in a large one-sided result. Finally, the sum of the square errors for content and style ($E_{PSNR}, E_{SSIM}$) is presented as the final indicator of the performance evaluation.

\begin{equation}
    w_{avg} = \frac{\sum_{i=1}^{5}{x_i w_i}}{\sum_{i=1}^{5}{w_i}} 
\end{equation}

\begin{equation}
    E_d = (w_{avg} - x_i)^2, \quad d = \{content, style\} 
\end{equation}

Tab. 3 and 4 show the results of the proposed performance indicators based on PSNR and SSIM, which were evaluated based on 1,452 generated results. PSNR values for content and style images are represented by $P_Content$ and $P_Style$ respectively, whereas $S_Content$ and $S_Style$ calculate the SSIM values for content and style images, respectively. $E_Content+E_Style$, which is the sum of the square errors for content and style, is used as the final metric. The preservation of content was highest when $L_h$ was not used, while not using $L_c$ or $L_s$ resulted in a loss of content and style. $L_l$ showed no significant difference in terms of content, whereas style was relatively high. Therefore, $E_PSNR$ and $E_SSIM$ are good evaluation metrics when all loss functions are used. The distribution of the results generated with the content retention performance as the style transfer performance is shown in Fig. 12. When PSNR is considered, the distribution of results with respect to $L_h$ is different from the other results. The distribution with respect to $L_c$ is located in the first half and with respect to $L_s$ in the second half. However, the distribution of $L_Total$ is relatively close to the center with a small deviation, making it the most appropriate result. In the case of SSIM, the distribution shape is similar to that of the PSNR, but several distributions show parallel movement results. The smaller the $E_SSIM$, the more central the distribution, indicating better performance. Therefore, $L_Total$ shows better performance than $L_l$, and $L_Total$ with a small difference has similar distributions in the center. Other results are considered to produce relatively poor results because they are located away from the center.

\begin{table}[htbp]
    \caption{Ablation Study Analysis (PSNR)}
        \centering   
        \begin{tabular}{cccccc}
            \hline
             & \multicolumn{5}{c}{PSNR} \\
            \cline{2-6} 
             & w.o $L_c$ & w.o $L_s$ & w.o $L_l$ & w.o $L_h$ & $L_{Total}$ \\
            \hline
            $P_{Content}$ & 8.690 & 9.833 & 9.242 & 17.350 & 9.280 \\
            $P_{Style}$ & 19.812 & 14.970 & 18.568 & 9.791 & 17.642 \\
            $E_{Content}$ & 1.745 & 0.032 & 0.591 & 53.868 & 0.534\\
            $E_{Style}$ & 9.045 & 3.368 & 3.109 & 49.301 & 0.701 \\
            \midrule
            $E_{PSNR}$ & 10.789 & 3.399 & 3.699 & 103.069 & \textbf{1.235}\\
            \hline
    \end{tabular}
\end{table}

\begin{table}[htbp]
    \caption{Ablation Study Analysis (SSIM)}
        \centering   
        \begin{tabular}{cccccc}
            \hline
             & \multicolumn{5}{c}{SSIM} \\
            \cline{2-6} 
             & w.o $L_c$ & w.o $L_s$ & w.o $L_l$ & w.o $L_h$ & $L_{Total}$ \\
            \hline
            $P_{Content}$ & 0.394 & 0.599 & 0.517 & 0.772 & 0.504 \\
            $P_{Style}$ & 0.715 & 0.466 & 0.599 & 0.338 & 0.597 \\
            $E_{Content}$ & 0.022 & 0.003 & 0.001 & 0.053 & 0.001 \\
            $E_{Style}$ & 0.025 & 0.009 & 0.002 & 0.049 & 0.001 \\
            \midrule
            $E_{PSNR}$  & 0.047 & 0.012 & 0.003 & 0.101 & \textbf{0.002} \\
            \hline
    \end{tabular}
\end{table}

\begin{figure}[htb!]
    \centering
    \includegraphics[width=14cm]{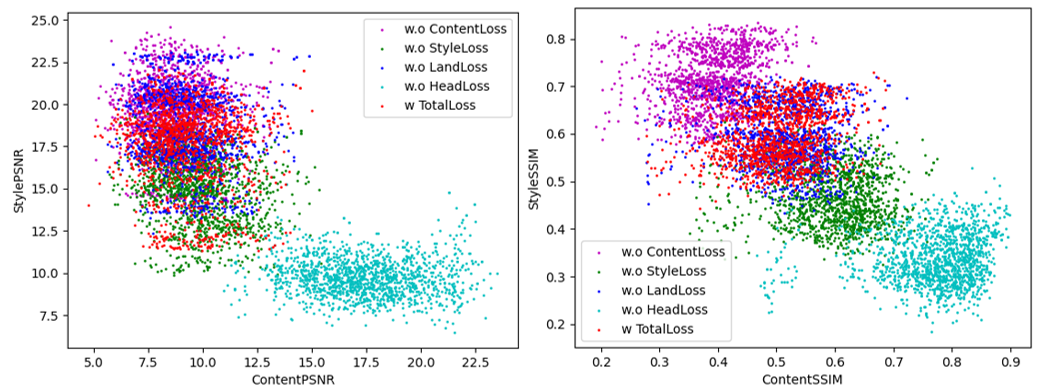}
    \caption{The results of content and style represent points for the test datasets through a two-dimensional coordinate system (top: PSNR, bottom: SSIM)}
\end{figure}

\section{Conclusions}
The objective of this study is to propose a generative adversarial network that utilizes facial feature points and loss functions to achieve arbitrary style transfers while maintaining the original face shape and transferring the Gat. To preserve the characteristics of the face, two loss functions, Land Loss and Head Loss, were defined using landmark masks to minimize the difference and speed up the learning process. Style Loss, which uses a gram matrix for content loss and style transfer, enables style transfer while preserving the character's shape. However, if the input images have large differences and the feature maps have significant discrepancies, the results are not satisfactory, and there are color differences in some instances. To overcome these limitations, it is recommended to define a loss function that considers color differences and aligns the feature map through the alignment of facial landmarks in future studies.

\bibliographystyle{unsrt}

\documentclass{article}


\usepackage{arxiv}

\usepackage[utf8]{inputenc} 
\usepackage[T1]{fontenc}    
\usepackage{hyperref}       
\usepackage{url}            
\usepackage{booktabs}       
\usepackage{amsfonts}       
\usepackage{nicefrac}       
\usepackage{microtype}      
\usepackage{lipsum}
\usepackage{graphicx}
\graphicspath{ {./images/} }


\title{PP-GAN : Style Transfer from Korean Portraits to ID Photos Using Landmark Extractor with GAN}


\author{
 Jongwook Si \\
  Dept. Computer AI Convergence Engineering\\
  Kumoh National Institute of Technology\\
  Gumi, KOREA 39177 \\
  \texttt{jwsi425@kumoh.ac.kr} \\
   \And
 Sungyoung Kim \\
  Dept. Computer Engineering\\
  Kumoh National Institute of Technology\\
  Gumi, KOREA 39177 \\
  \texttt{sykim@kumoh.ac.kr} \\
  \And
}

\begin{document}
\maketitle
\begin{abstract}
The objective of a style transfer is to maintain the content of an image while transferring the style of another image. However, conventional research on style transfer has a significant limitation in preserving facial landmarks, such as the eyes, nose, and mouth, which are crucial for maintaining the identity of the image. In Korean portraits, the majority of individuals wear "Gat", a type of headdress exclusively worn by men. Owing to its distinct characteristics from the hair in ID photos, transferring the "Gat" is challenging. To address this issue, this study proposes a deep learning network that can perform style transfer, including the "Gat", while preserving the identity of the face. Unlike existing style transfer approaches, the proposed method aims to preserve texture, costume, and the "Gat" on the style image. The Generative Adversarial Network forms the backbone of the proposed network. The color, texture, and intensity were extracted differently based on the characteristics of each block and layer of the pre-trained VGG-16, and only the necessary elements during training were preserved using a facial landmark mask. The head area was presented using the eyebrow area to transfer the "Gat". Furthermore, the identity of the face was retained, and style correlation was considered based on the Gram matrix. The proposed approach demonstrated superior transfer and preservation performance compared to previous studies. 
\end{abstract}




\section{Introduction}
With the advent of modern technologies, such as photography, capturing the appearances of people has become effortless. However, when these technologies were not developed artists would paint portraits of individuals. Such a painting is called a portrait, and because of the invention of photography, modern portraits have become a new field of art. However, all famous figures from the past were handed down in pictures. The main purpose of paintings is to depict politically famous figures, but in modern times, the purpose has expanded to the general public. Although the characteristics of portraits by period and country are very different, most differ greatly from the actual appearance of the characters unless they are surrealistic works. Korean portraits differ considerably depending on time and region. Fig. 1(a) shows a representative work of portraits from the Goryeo Dynasty. This work is a portrait of Ahn Hyang, a Neo-Confucian scholar from the mid-Goryeo period. Fig. 1(b) is a portrait of the late Joseon Dynasty, which indicates that there is a large difference in the preservation conditions and drawing techniques. In particular, in Fig. 1(b), the "Gat" on the head is clearly visible.

\begin{figure}[htb!]
    \centering
    \includegraphics[width=10cm]{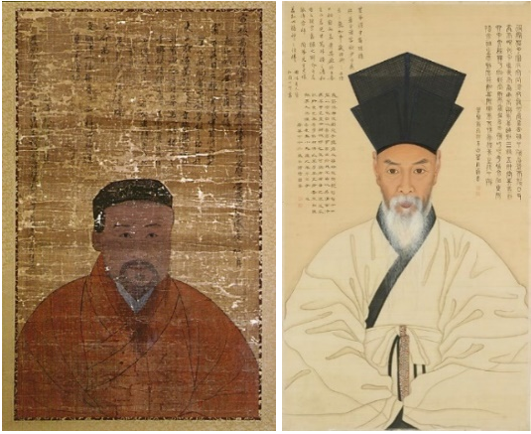}
    \caption{The left photo (a) is a portrait of Hyang An (1243 ~ 1306) in the mid-Goryeo dynasty and the right photo (b) is a portrait of Chae Lee (1411 ~ 1493) in the late Joseon Dynasty}
\end{figure}

Prior to the Three Kingdoms Period, Korean portrait records were absent, and only a limited quantity of portraits were preserved during the Goryeo Dynasty [1]. In contrast, the Joseon Dynasty produced numerous portraits with different types delineated according to their social status. Furthermore, works from the Joseon era exhibit a superior level of painting, in which facial features are rendered in a greater detail than in earlier periods.
A portrait exhibits slight variations in the physical appearance of a person, but it uniquely distinguishes individuals akin to a montage. Modern identification photographs serve a similar purpose and are used as identification cards, such as driver's licenses and resident registration cards. Old portraits may pique interest in how one appears in such artwork, for which style transfer technology can be used. Korean portraits may be used to provide the style of ID photos; however, the custom of wearing a “Gat” headgear renders transferring the style from Korean portraits to ID photos using previous techniques challenging. While earlier studies have employed global styles or partial styles to transfer onto content images, the distinct styles of texture, attire, and “Gat” must be considered simultaneously for Korean portraits. By independently extracting several styles from the style image, transferring the age, hairstyle, and costume of the person in a portrait onto the ID photo is possible. Fig. 2 depicts the outcome of style transfer using CycleGAN, a prevalent method for style transfer, revealing the difficulty of achieving adequate style transfer when numerous styles are involved. In this study, we introduce a method for high-quality style transfer of Korean portraits, which overcomes the limitations of previous research to accurately preserve facial landmarks and produce realistic results.

\begin{figure}[htb!]
    \centering
    \includegraphics[width=12cm]{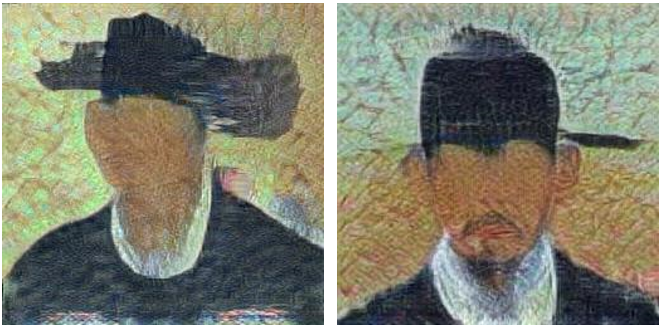}
    \caption{Results of style transfer from Korean portraits to ID photos using CycleGAN}
\end{figure}

Style transfer techniques, such as GAN, are commonly used based on facial datasets, but maintaining the identity of the person is crucial for achieving high-quality results. Existing face-based style transfer studies only consider facial components, such as eyes, nose, mouth, and hair, when transferring styles onto content images. In contrast, this study aims to transfer multiple styles, including Gats and costumes, simultaneously. To accomplish this, we propose an enhanced GAN-based network for style transfer that generates a mask using landmarks and defines a new loss function to perform style transfer based on facial data. we define the proposed method, "Style Transfer from Korean Portraits to ID Photos Using Landmark Extractor with GAN" as PP-GAN. The primary contribution of this study is the development of a novel approach to style transfer that considers multiple styles and maintains the identity of a person.

•	The possibility of independent and arbitrary style transfer to a network trained with a small dataset has been demonstrated. \\
•	This study is the first attempt at arbitrary style transfer in Korean portraits, which was achieved by introducing a new combination of loss functions.\\
•	The generated landmark mask improved the performance of identity preservation and outperformed previous methods [2].\\
•	New data on upper-body Korean portraits and ID photo were collected for this study.\\



\section{Related Works}
\label{sec:headings}
Research on style transfer can be categorized into two main groups: those based on convolutional neural Networks [4-6], [8-13] and those based on general adversarial networks [3], [14-23].

\subsection{CNN-Based Style Transfer}
\label{sec:headings}
AdaIN [4] suggested a method of transferring style at high speed using statistics in feature maps of content and style images. This is one of the earlier studies on style transfers. S. Huang et. al. [5] used the correlation between the content feature map and scaling information of the style feature map for the fusion of content and style. In addition, the order statistics method, called “Style Projection”, demonstrated the advantages and results of fast training speed. Zhu et. al. [6] maintained structural distortion and content by presenting a style transfer network that could preserve details. In addition, by presenting the refined network, which modified the VGG-16[7], the style pattern was preserved via spatial matching of hierarchical structures. Elad et. al. [8] proposed a new style transfer algorithm that expanded the texture synthesis work. It aimed to create images of similar quality and emphasized a consistent way of creating rich styles while keeping the content intact in the selected area. In addition, it was fast and flexible to process any pair of content and style images. S. Li et. al. [9] suggested a style transfer method for low-level features to express content images in a CNN. Low-level features dominate the detailed structure of new images. A Laplacian matrix was used to detect edges and contours. It shows a better stylized image, which can preserve the details of the content image and remove artifacts. Chen et. al. [10] proposed a stepwise method based on a deep neural network for synthesizing facial sketches. It showed better performance by proposing a pyramid column feature to enrich the parts by adding texture and shading. Fast Art-CNN [11] is a structure for fast style transfer performance in the feedforward mode while minimizing deterioration in image quality. It can be used in real-time environments as a method for training deconvolutional neural networks to apply a specific style to content images. X. Lio et. al. [12] proposed an architecture that includes geometric elements in the style transfer. This new architecture can transfer textures into distorted images. In addition, because the content/texture-style/geometry style can be selected to be entered in triple, it provides much greater versatility to the output. P. Kaur et. al. [13] proposed a framework that solves the problem of realistically transferring the texture of the face from the style image to the content image without changing the identity of the original content image. Changes around the landmark are gently suppressed to preserve the facial structure so that it can be transferred without changing the identity of the face. 

\subsection{GAN-Based Style Transfer}
\label{sec:headings}
APDrawingGAN [14] improved the performance by combining global and regional networks. High-quality results were generated by measuring the similarity between the distance transform and artist drawing. Zheng Xu et. al. [15] used a generator and discriminator as conditional networks. Subsequently, the mask module for style adjustment and AdaIN [4] for style transfer performed better than existing GAN. S3-GAN [16] introduced a style separation method in the latent vector space to separate style and content. A style-transferred vector space was created using a combination of separated latent vectors. CycleGAN [3] proposes a method for converting a style to an image without a pair of domains. While training the generator mapping to X → Y, reverse mapping to Y → X is performed. In addition, the cycle consistency loss was designed such that an input image and its reconstructed image could be identical when the transferred style was removed through reverse mapping. In SLGAN [17], a style-invariant decoder was created by a generator to preserve the identity of the content image and introduce a new perceptual makeup loss, resulting in high-quality conversion. Some attempts have been made to maintain facial landmarks in style transfer studies aimed at makeup [18-20] or aging [21-22]. BeautyGAN [18] defines instance and perceptual loss to change the makeup style while maintaining the identity of the face, thereby generating high-quality images and maintaining the identity. Paired-CycleGAN [19] trains two generators simultaneously to convert the makeup styles of other people from portrait photos. Stage 1 was used as a pair of powers through image analogy, and as an input of Stage 2, it showed excellent results by calculating identity preservation and style consistency compared to the power of Stage 1. Landmark-CycleGAN [20] showed incorrect results owing to the distortion of the geometrical structure while converting a face image to a cartoon image. To solve this problem, local discriminators have been proposed using landmarks to improve performance. Palsson's et. al. [21] suggested Group-GAN, which consisted of several models of CycleGAN[3], to integrate pre-trained age prediction models and solve the face aging problem. Wang et. al. [22] proposed a method for interconverting edge maps to a CycleGAN-based E2E-CycleGAN network for aging. The old face was generated using the identity feature map and result of converting the edge map using the E2F-pixelHD network. Yi et. al. [23] proposed a new asymmetric cycle mapping that forced the reconstruction information to be shown and included only in optional facial areas. Portrait images generated along with a localized discriminator for landmark and style classifiers were introduced. Considering the style vector, portraits were generated in several styles using a single network. They attempted to transfer the style of the portrait similar, which is similar to the purpose of our study. However, in this study, not only the portrait painting style but also the Gat and costume are transferred together.


\section{Background}
\subsection{VGG-16}
\label{sec:headings}
The VGG-16 [7] network is a prominent computer vision model that attained a 92.7\% Top-5 accuracy in the ImageNet Challenge competition by receiving an RGB image with dimensions of 224 × 224 as input, containing 16 layers in a configuration of 13 convolution layers and three FC layers. The convolution filter measures 3 × 3 pixels and maintains fixed strides and padding at 1. The activation function employed in the network is ReLU, and the pooling layer is max pooling, which is set to a fixed stride of 2 on 2 × 2. The closer it is to the input layer, the more low-level information the feature map contains, such as color and texture. of the image, and the closer it is to the output layer, thus providing higher-level information, such as shape. The pre-trained VGG-16 [7] is used in this study to preserve facial and upper-body content and transfer the style efficiently.

\subsection{Gram matrix}
\label{sec:headings}
The Gram matrix is a valuable tool for representing the color distribution of an image. This enabled the computation of the overall color and texture correlation between the two images. Leon et. al.[24] demonstrated that the style transfer performance can be improved using a Gram matrix on feature maps from various layers. Fig. 3 illustrates the process of calculating the Gram matrix, which involves converting each channel of a color image into a 1D vector, followed by obtaining the matrix by multiplying the H×W matrix with its transpose. The Gram matrix is a square matrix with channel size as its dimension. As the corresponding values in the Gram matrix of the two images become more similar, the color distribution of the images also becomes more similar.

\begin{figure}[htb!]
    \centering
    \includegraphics[width=13cm]{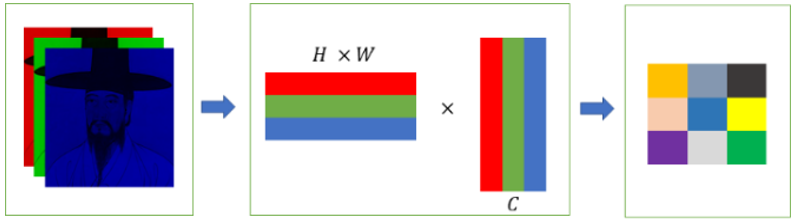}
    \caption{The process of calculating into a gram matrix for Korean portraits}
\end{figure}


\subsection{Face Landmark}
\label{sec:headings}
Facial landmarks, such as the eyes, nose, and mouth, play a vital role in identifying and analyzing facial structures. To detect the landmarks, this study employed the Shape Predictor of 68 face landmarks [25], which generated 68 x and y coordinates of the crucial facial components, including the jaw, eye, eyebrow, nose, and mouth and also provided the locations of the face. Subsequently, the coordinates obtained from the predictor were used to create masks for the eyes, nose, and mouth, as shown in Fig. 4.

\begin{figure}[htb!]
    \centering
    \includegraphics[width=10cm]{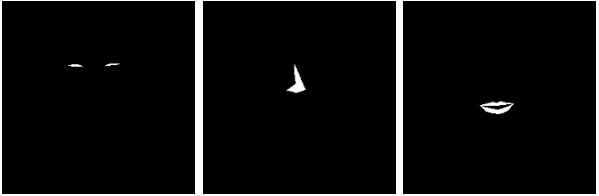}
    \caption{Masks for the eyes, nose, and mouth created using the coordinates returned by Shape Predictor 68 Face Landmarks}
\end{figure}


\subsection{Image Sharpening}
\label{sec:headings}
Image sharpening is considered a high-frequency emphasis filtering technique, which is employed to enhance image details. High frequency is characterized by the changes in brightness or color occurring locally, and it is useful in identifying facial landmarks. Image sharpening can be achieved using high-boost filtering. It involves generating a high-pass image by subtracting a low-pass image from an input image as shown in Eq. (1). A high-frequency emphasized image is obtained by multiplying the input image with a constant during this process.

\begin{equation}
g(x,y) = Af(x,y) - f_L(x,y) 
\end{equation}

Mean filtering is a low-pass filtering technique, and the coefficients of the filter can be determined using Eq. (2). The sharpening strength of the input image is controlled by the value of $\alpha$, where $9A-1$ is set to $\alpha$. A high $\alpha$ value results in a decrease in the sharpness level owing to the high ratio of the original image to the output. Conversely, a small $\alpha$ value results in a reduction in contrast, owing to the removal of numerous low-frequency components. 
To have similar structures between portrait images and ID photos, portrait images are cropped around the faces, as the face occupies a relatively small area. In contrast, ID photos are resized so that they have the same size, both horizontally and vertically, instead of being cropped. However, this resizing can make extracting facial landmarks difficult. Therefore, image sharpening is performed in the present study. This process is necessary to ensure that facial landmarks are extracted well from ID photos, as shown in Fig. 5, where the difference in facial landmark extraction with and without image sharpening is illustrated.

\begin{equation}
    A
    \left[
    \begin{array}{ccc}
        0 & 0 & 0 \\
        0 & 1 & 0 \\
        0 & 0 & 0 \\
    \end{array}
    \right]
    - 1/9
    \left[
    \begin{array}{ccc}
        1 & 1 & 1 \\
        1 & 1 & 1 \\
        1 & 1 & 1 \\
    \end{array}
    \right]
   = 1/9
    \left[
    \begin{array}{ccc}
        -1 & -1 & -1 \\
        -1 & 9A-1 & -1 \\
        -1 & -1 & -1 \\
    \end{array}
    \right]
   \rightarrow
   \left[
    \begin{array}{ccc}
        -1 & -1 & -1 \\
        -1 & \alpha & -1 \\
        -1 & -1 & -1 \\
    \end{array}
    \right]
\end{equation}

\begin{figure}[htb!]
    \centering
    \includegraphics[width=13cm]{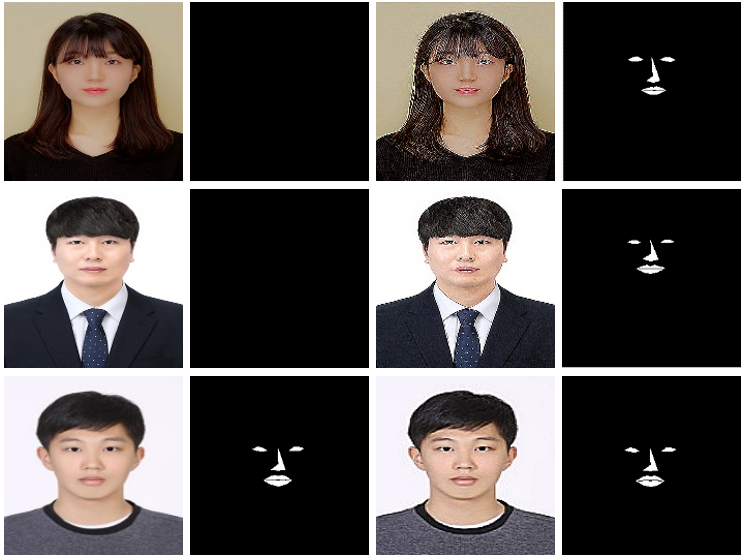}
    \caption{Result of landmark mask generation according to the use of high boost filtering (The first and third columns are the original and the high boost filtered image, respectively, and the second and fourth columns show the masks with the detected landmark for each corresponding image)}
\end{figure}


\section{Proposed Method}
\label{sec:headings}

\subsection{Network}
\label{sec:headings}

The primary objective of the proposed method is to achieve a style transfer of ID photos to Korean portraits. Let $X$ and $Y$ indicate the domains of the three-dimensional color image and the Korean portrait, respectively. These domains are subsets of $X \subseteq \mathbb{R}^{H \times W \times C}$ and $Y \subseteq \mathbb{R}^{H \times W \times C}$ and have a set relationship such that $x \in X$ and $y \in Y$.

The CycleGAN [3] network is limited in performing style transfer owing to its training over the entire domain. Therefore, the proposed method adopts a Dual I/O generator from BeautyGAN [18], which has a stable discriminator that enables mapping training between two domains and style transfer. Additionally, the proposed method incorporates VGG-16 ,a gram matrix, and a landmark extractor to improve performance. Fig. 6 depicts the overall structure of the proposed method.



\subsubsection{Generator}

The generator is trained to perform $(X, Y) \rightarrow (Y, X)$ mapping, resulting in a fake image $G(x, y) = (x_y, y_x)$ with content $X$ and style $Y$, which is evaluated in this study. Contents $Y$ and $X$ are used to generate another fake image $y_x$. This study focuses only on $x_y$ results, even though the network structure can generate results in both directions.
The image recovered by the Dual I/O generator performing style transfer and the input image must be identical. With an input size of $(256, 256, 32)$, $x$ and $y$ pass through three convolution layers each, resulting in a size of $(64, 64, 128)$. The $x$ and $y$ results are concatenated to produce a size of $(64, 64, 256)$, which is restored to the original size through the deconvolution layer, allowing style transfer through nine residual blocks. This result represents a fake image of a style transferred and represents the result of the proposed method.
Therefore, the generator deceives the discriminator by generating fake images that appear real, resulting in more natural and higher-performance results.


\subsubsection{Discriminator}

The network structure includes two discriminators that are trained to classify the styles of fake and real images generated by the generator. The discriminator consists of five convolution layers and aims to distinguish styles. The input image size is $(256, 256, 3)$, and the network result size is $(30, 30, 1)$. The first four convolution layers, excluding the last layer, perform Spectral Normalization [26] to improve performance and maintain a stable distribution of the discriminator in a high-dimensional space.
The discriminator is defined as follows: $D_x$ classifies $x_y$ as fake and $y$ as real, whereas $D_y$ classifies $y_x$ as fake and $x$ as real. Finally, the PatchGAN [27] is applied to produce the discriminator output, which is the final judgment result of the discriminator on the input image.

\begin{figure}[htb!]
    \centering
    \includegraphics[width=16cm]{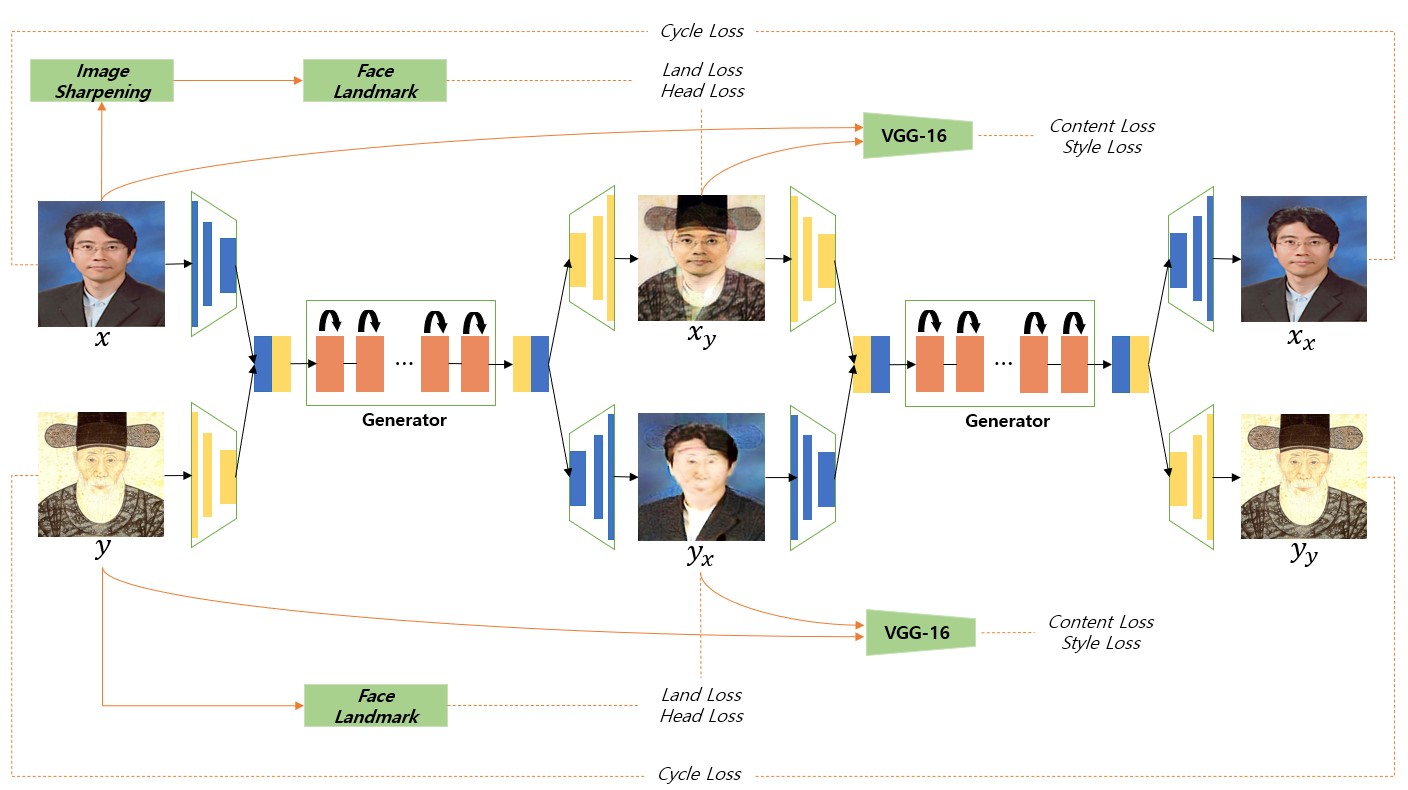}
    \caption{Overall structure of the system proposed in this study}
\end{figure}



\subsection{Loss Functions}
\label{sec:headings}
In this study, we propose a loss function for transferring ID photos to arbitrary Korean portrait styles. Six loss functions, including the loss function of the new approach, are used to generate good results.
CycleGAN introduced the concept of setting the result as the input of the generator again through a cycle structure, which should theoretically produce the same output as the original image. Therefore, in this study, we define the recovered result as the cycle loss, which consists of a loss function designed to reduce the difference between the input and output images. In particular, it can be expressed as $x \approx G(G(x,y)) = G(x_y, y_x) = x_x$ and $y \approx G(G(y,x)) = G(y_x, x_y) = y_y$. This can be expressed as Eq. (3).\\

\begin{equation}
L_{cy} = \mathbb{E}_{x \sim P(X)} \|x_x - x\| + \mathbb{E}_{y \sim P(Y)} \|y_y - y\}
\end{equation}

The existing style transfer method distorts the shape of the face geometrically, leading to difficulties in recognizing the face shape. To maintain the identity of the character, a new condition is required. Hence, this study defines land loss based on a face landmark mask, which helps in preserving the eyes, nose, and mouth while enhancing the performance of style transfer. Land loss is defined by the mathematical expression Eq. (4) in this study.

\begin{equation}
L_l = L_{l_{eye}} + L_{l_{nose}} + L_{l_{lip}}
\end{equation}


Land loss is a function that aims to maintain the landmark features of the input and output images generated by the generator. The pairs of images $(x_y, x)$ and $(y_x, y)$ contain the same content with different styles, and the landmark shapes are identical. The masks $M_{fX}$ and $M_{fY}$, generated for the eye, nose, and mouth areas, are used to calculate the area, as discussed in Section 3. Using a pixel-wise operation, each area of the eye, nose, and mouth is processed using a face landmark mask, and a loss function is defined to minimize the difference in pixel values. This process is expressed in Eq. (5). The difference for each landmark is based on L1 Loss.

\begin{equation}
L_f = \mathbb{E}_{x \sim P(X)} \|x_y \odot M_{fX} - x \odot M_{fX}\|_1 + \mathbb{E}_{y \sim P(Y)} \|y_x \odot M_{fY} - y \odot M_{fY}\|_1, \quad f = \{l_{eye}, l_{nose}, l_{lip}\}
\end{equation}


The method proposed in this study differs greatly from previous style transfer research, which requires some content of the style target image rather than ignoring it and only considering the color relationship. In particular, for Korean portraits, the style of the Gat and clothes must be considered in addition to image quality, background, and overall color. However, the form of the Gat varies widely, which is difficult to detect due to the differences in wearing position, while the hair of Korean portraits and ID photos have completely different shapes. To address this, a head loss is proposed to minimize the difference between the head area of the result and style images, with the head area divided into the Gat and hair areas, represented by masks $M_{ht}$ and $M_{hr}$. Head loss uses the fact that the Gat does not cover the eyebrows; therefore, the feature point located at the top of the coordinates corresponding to the eyebrows is used to define the head area, which is then used to transfer the corresponding style to the resulting image. This is expressed in Eq. (6).


\begin{equation}
L_h = \mathbb{E}_{x \sim P(X)} \|x_y \odot M_{ht} - y \odot M_{ht}\|_1 + \mathbb{E}_{y \sim P(Y)} \|y_x \odot M_{hr} - x \odot M_{hr}\|_1
\end{equation}



To preserve the overall shape of the character and enhance the performance of style transfer, content loss and style loss are defined using a specific layer of VGG-16 in this study. The pre-trained network contains low- and high-level information, such as colors and shapes, which appear differently depending on the layer location. Low-level information is related to style, and high-level information is related to content. Conversely, high-level layers represent the image characteristics. Therefore, the content and style losses are configured based on the layer characteristics. Style loss is defined using a gram matrix, which is obtained by computing the inner product of the feature maps. The best set of layers obtained through the experiment is used to define style loss, as shown in Eq. (7), where $N$ and $M$ represent the product and channel of each layer, respectively, and $g$ represents the gram matrix of the feature map. By training to minimize the difference in the gram matrix between the feature maps for both sides ($x_y$ and $y_x$), the style of $y$ can be transferred to $x$.

\begin{equation}
L_s = \frac{1}{4N^2M^2} \sum \left[ \left(g_i(x_y) - g_i(y)\right)^2 + \left(g_i(y_x) - g_i(x)\right)^2 \right]
\end{equation}



Content loss is defined as a method to minimize linear differences in feature maps at the pixel level. Because the style transfer aims to maintain the content of an image while transferring the style, it is not necessary to consider correlations. The equation for content loss is the same as that in Eq. (8). This is a critical factor in preserving the identity of a person; however, if the weight of this loss is extremely large, it can result in poor style transfer results. Therefore, appropriate hyperparameters must be selected to achieve the desired outcome.

\begin{equation}
L_c = \mathbb{E}_{x \sim P(X)} [l_i(x_y) - l_i(x)]^2 + \mathbb{E}_{y \sim P(Y)} [l_i(y_x) - l_i(y)]^2
\end{equation}



The generator loss is composed of cycle, land, head, style, and content losses, as expressed in Eq. (9). Each loss is multiplied by a different hyperparameter, and the sum of the resulting values is used as the loss function of the generator.

\begin{equation}
L_G = \lambda_{cy} L_{cy} + \lambda_{l} L_{l} + \lambda_{h} L_{h} + \lambda_{s} L_{s} + \lambda_{c} L_{c} 
\end{equation}



The discriminator loss solely comprises adversarial loss, which follows the GAN structure. The output of the discriminator is a $32 \times 32 \times 1$ result that is evaluated based on PatchGAN [27] to identify whether they are authentic or fake, considering every image PatchGAN [27]. The loss function used to train the discriminator is given by Eq. (10). The loss function is reduced if the patches of $x_y$ and $y_x$ are fake, and the patches of $x$ and $y$ are genuinely classified.

\begin{equation}
L_D = \mathbb{E}_{x \sim P(X)} [(D_x(y) - 1)^2 + (D_x(x_y))^2] + \mathbb{E}_{y \sim P(Y)} [(D_y(x) - 1)^2 + (D_y(y_x))^2]
\end{equation}



The total loss employed in this study is expressed by Eq. (11) and is composed of the generator and discriminator losses. The generator seeks to minimize the generator loss to generate style transfer outcomes, whereas the discriminator aims to minimize the discriminator loss to enhance its discriminative capability. A trade-off between the generator and discriminator performances is observed, where if one is improved, the other is diminished. Consequently, the total loss is optimized by forming a competitive relationship between the generator and discriminator, which leads to superior outcomes.


\begin{equation}
L_{Total} = \min_G \min_D (L_G + L_D)
\end{equation}


\subsection{Training}
The experimental environment in this study was conducted on a multi-GPU system using the GeForce RTX 3090 and Ubuntu 18.04 LTS operating system. As TensorFlow 1.x has a minimum version requirement for CUDA, the experiments were carried out using Nvidia-Tensorflow version 1.15.4.
Datasets of ID photos and Korean portraits were collected through web crawling using Google and Bing search engines. To improve the training performance, preprocessing was conducted to separate the face area from the whole body of the Korean portraits, which typically feature the entire body. Data augmentation techniques, such as left and right inversion, blur, and noise, were applied to increase the limited number of datasets. Gat preprocessing was also performed, as shown in Fig. 7, to facilitate the feature mapping.
Tab. 1 shows the resulting dataset consisting of 1,054 ID photos and 1,736 Korean portraits divided into 96\% training and 4\% test sets. Owing to the limited number of portraits, a higher ratio of training data was used, and no data augmentation was applied to the test set. As the number of combinations that could be generated from the test data was substantial ($X_{Test}$ × $Y_{Test}$), the evaluation was not problematic.
Previous research has emphasized the importance of data preprocessing, and the results of this study further support its impact on training performance.

\begin{figure}[htb!]
    \centering
    \includegraphics[width=9cm]{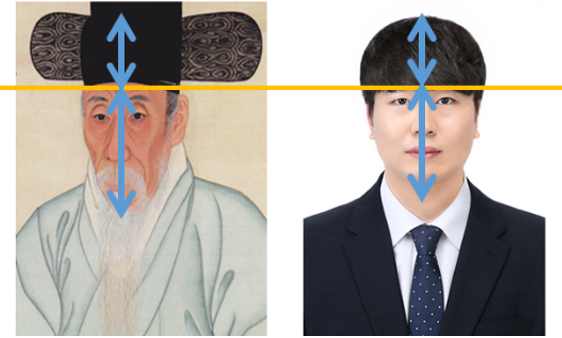}
    \caption{Examples of datasets preprocessing}
\end{figure}

\begin{table}[htbp]
 \caption{Detailed Datasets}
  \centering
  \begin{tabular}{cccc}
    \toprule
    Data     & Train     & Test & Sum \\
     \midrule
    ID Photos($X$) & 988 & 66 & 1,024\\

    Korean Portraits($Y$) & 1,714 & 22 & 1,736\\
     \midrule
     Sum($X+Y$) & 2,702 & 88 & 2,790\\
     \midrule   
     Combination($XY$) & 1,693,432 & 1,452 & -\\
    \bottomrule
  \end{tabular}
  \label{tab:table}
\end{table}

The proposed network was trained for 200 epochs using the Adam Optimizer. The initial learning rate was set to 0.0001 and linearly reduced to zero after 50\% of the training epoch for stable learning. To match the equality between the loss functions, \(\lambda_{cy}\) was set to 50, which resulted in a relatively lower value than the other losses. To increase the effect of style transfer, \(\lambda_s\) was set to 1 and \(\lambda_h\) was set to 0.5, which helped concentrate on the head area between the style transfers. Finally, the training proceeded by setting \(\lambda_c = 0.1\) and \(\lambda_l = 0.2\). The entire training process took approximately 6.5 hours. The results are presented in Fig. 8, which visually confirms that the proposed method shows a greater performance improvement than previous research [2]. While previous methods have focused only on style transfer, this study successfully maintained the identity of a person while transferring the style. The results show excellent outcomes in which the style is transferred while preserving the shape of the character in the content image. Additionally, the identity of the personnel is preserved, and the Gat is transferred naturally.


\begin{figure}[htb!]
    \centering
    \includegraphics[width=16cm]{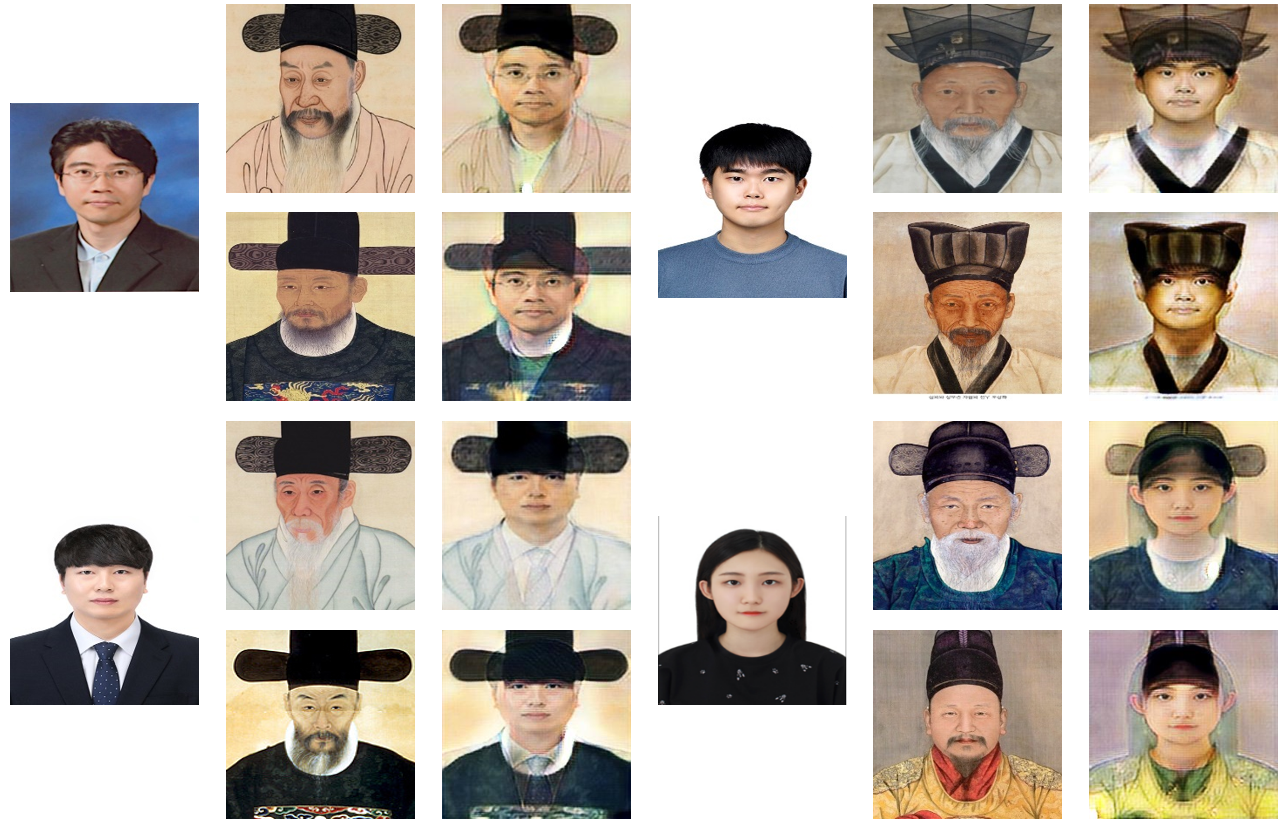}
    \caption{The result of the method proposed in this paper}
\end{figure}

\section{Experiments}
\label{sec:headings}

\subsection{Feature Map}
\label{sec:headings}
To perform style loss, this study adopted Conv2\_2 and Conv3\_2 from the VGG-16 layer, whereas Conv4\_1 was used for content loss. Although the front convolution layers contain low-level information, they are sensitive to change and difficult to train because of the large differences in pattern and color between the style and content images. To overcome this problem, this study utilizes a feature map located in the middle to extract low-level information for style transfer. The results of using a feature map not adopted for style loss are presented in Fig. 9 with a smoothing set to 0.8 on the tensorboard. Loss graphs were output for only ten epochs because of the failure of training when layers not used for Style Loss were utilized in the experiment.

•	The Conv2\_1 layer exhibits a large loss value and unstable behavior during training, indicating that training may not be effective for this layer.\\
•	Conv1\_2 is close to zero in most of the losses, but it cannot be said that the training proceeds because the maximum minimum differs by more than $10^4$ times owing to the very large loss of some data.\\
•	Conv1\_1 exhibits a high loss deviation and instability during training, similar to Conv2\_1 and Conv1\_2. Moreover, owing to its sensitivity to color, this layer presents challenges for training.


If Conv4\_1 is utilized as the style loss layer, it can transfer the style of the image content. However, because the feature map scarcely includes style-related content, the generator may produce images lacking style. Nevertheless, it is observed that transferring the style of the background is feasible as it corresponds to the overall style of the image and can be recognized as content because clothing style is not a prominent feature. Therefore, high-level layers, such as Conv4\_1, contain only the style of the background in the character content. The result of utilizing the feature map employed in the content loss for style loss is shown in Fig. 10. In general, most contents of the content image are preserved, whereas the style is marginally transferred. Hence, we proceed with using trainable layers, which results in stable training and enables us to transfer styles while conserving content.

\begin{figure}[htb!]
    \centering
    \includegraphics[width=15cm]{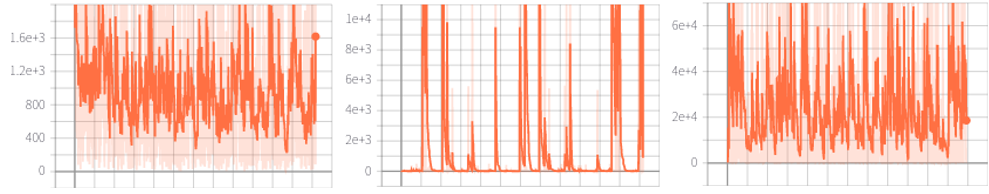}
    \caption{Result of a specific feature map experiment of VGG-16 for style loss}
\end{figure}

\begin{figure}[htb!]
    \centering
    \includegraphics[width=10cm]{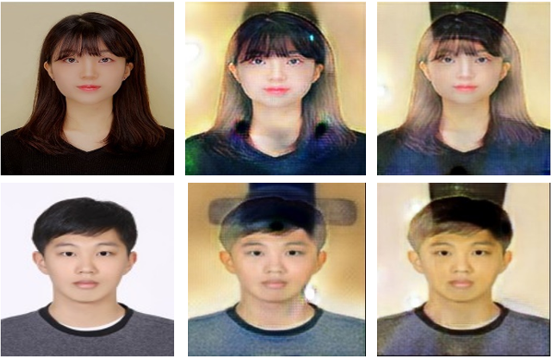}
    \caption{Results when the feature map used for Style Loss is used with the same layer as Content Loss (Column 1: Input Image; Column 2: and Column 3: output image using Conv4\_1)}
\end{figure}

\subsection{Ablation Study}
\label{sec:headings}
An ablation study was conducted on four loss functions, excluding Cycle Loss, to demonstrate the effectiveness of the loss function proposed in this study for Generator Loss. The results are presented in Fig. 11, where one row represents the use of the same content and style images. If $L_c$ is excluded, the character shape is not preserved, leading to poor results owing to the style concentration during training. Consequently, only facial components are transferred based on $L_l$. When $L_c$ and $L_l$ are excluded simultaneously, the style transfer outcome lacks facial components. Similarly, when $L_s$ is excluded, the result of the style transfer is of poor quality, with the character remaining almost the same. The use of $L_{cy}$ allows for style transfer of the background without the need for a separate loss function for style; however, owing to the main focus of training on the character shape, style transfer barely occurs, resulting in the creation of the Gat when $L_h$ is used. Excluding $L_l$ leads to unclear and blurred liver facial components, and the face color becomes bright. Therefore, $L_l$ plays a crucial role in preserving the character identity by making the liver face components more apparent. If $L_h$ is excluded, the head area becomes blurred or is not created, leading to unsatisfactory style transfer results. Unlike the overall style transfer, the Gat must be newly generated; thus, $L_s$ serves a different purpose. Therefore, the head area must be set separately, and $L_s$ can be used with $L_h$ to achieve this purpose. Consequently, the use of all the loss functions proposed in this study results in the best performance, generating natural images without bias in any direction.

\subsection{Performance Evaluation}
\label{sec:headings}
This study conducts a performance comparison with previous research [2] with the same subject, as well as an ablation study. Although there are diverse existing studies on style transfer, the subject of this paper differs significantly from them, leading to a comparison only with a single previous study [2].
Based on CycleGAN [3], evaluation in pairs could not be performed because of the impossibility of arbitrary style transfer. Thus, an evaluation survey was conducted with 59 students of different grades from the Department of Computer Engineering at Kumoh National Institute of Technology to evaluate the performance of the proposed method in terms of three items: the transfer of style ($S_{st}$), preservation of content ($S_{cn}$), and generation of natural images ($S_{nt}$). The survey was conducted online for 10 days using Google Forms, and the surveyors received and evaluated a combination of 10 results from a previous study [2] and 10 results from the proposed method. The survey results presented in Tab. 2 show that the proposed method outperforms the previous method [2] in all three aspects, with the highest difference in scores for preserving character content. In contrast, the previous method [2] failed to preserve the shape of the character during style transfer, leading to blurring or disappearance of the landmark of the face, which was not natural. However, the proposed method successfully preserved the character content and effectively transferred the style, producing relatively natural results. Thus, the proposed method showed better overall performance than the previous method [2].

\begin{table}[htbp]
 \caption{Survey Results}
  \centering
  \begin{tabular}{cccc}
    \toprule
    Score     & Previous method [2]    & Ours & Gap \\
     \midrule
    $S_{cn}$ & 2.510 & \textbf{3.867} & \underline{1.357}\\
    $S_{st}$ & 3.378 & \textbf{3.402} & 0.025\\
    $S_{nt}$ & 2.295 & \textbf{3.262} & 0.337\\
     \midrule
      Sum & 8.813 & \textbf{10.531} & 1.718\\
    \bottomrule
  \end{tabular}
  \label{tab:table}
\end{table}

\begin{figure}[htb!]
    \centering
    \includegraphics[width=15cm]{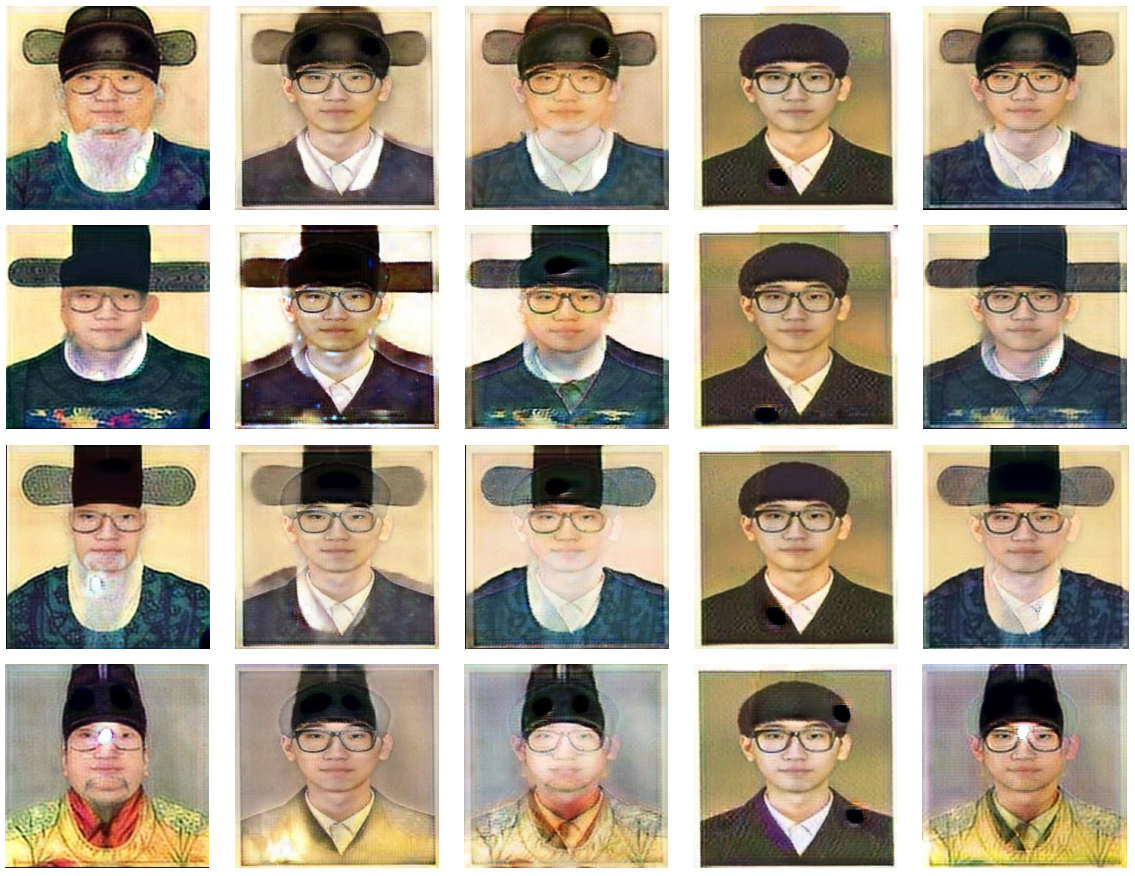}
    \caption{Results of the importance of the various loss functions that make up the Total Loss proposed in this paper. Column 1 to Column 4 Excluding the loss functions $L_c$, $L_s$, $L_l$, and $L_h$; Column 5 results generated using all loss functions}
\end{figure}


Peak signal-to-noise ratio (PSNR) and structural similarity index measure (SSIM) are commonly employed to measure performance, as observed in various studies [28-29]. However, when conducting style transfer while preserving content, it is crucial to ensure a natural outcome without any significant bias towards either the content or style. Consequently, to compare the performance, we propose new performance indicators that utilize a weighted arithmetic mean, which is further weighted by the median values of the PSNR and SSIM. The final result is obtained by combining the results of both content and style using the proposed performance indicator.
The PSNR is commonly used to assess the quality of images after compression by measuring the ratio of noise to the maximum value, which can be calculated using Eq. (12). The logarithmic denominator in this equation represents the average sum of squares between the original and compressed images, and a lower value indicates a higher PSNR and better preservation of the original image. By contrast, the SSIM is used to evaluate distortions in image similarity between a pair of images by comparing their structural, luminance, and contrast features. Eq. (13) is used to calculate the SSIM, which involves various probability-related definitions such as mean, standard deviation, and covariance.

\begin{equation}
    PSNR(A,B) = 10\log_{10}\left(\frac{MAX^2}{\sum{(A-B)^2}}\right) 
\end{equation}

\begin{equation}
    SSIM(A,B) = \frac{(2\mu_A\mu_B + C_1)(2\sigma_{AB} + C_2)}{(2\mu_A^2 + \mu_B^2 + C_1)(\sigma_A^2 + \sigma_B^2 + C_2)} 
\end{equation}

To evaluate the performance, the indicators are sorted in ascending order, resulting in a sequence of values denoted as $[x_1,x_2,x_3,x_4,x_5]$, where $x_3$ represents the best result. To assign more weight to the median value, a weight vector ($w$) of $[10, 25, 50, 25, 10]$ is assigned, and the weighted arithmetic mean is calculated using Eq. (14). The performance is evaluated using Eq. (15), which calculates the square of the difference between the weighted arithmetic mean and PSNR and SSIM values. The resulting value indicates the degree of performance, with smaller values indicating better performance. The difference from the average weight ($w_{avg}$) is squared and added, resulting in a large one-sided result. Finally, the sum of the square errors for content and style ($E_{PSNR}, E_{SSIM}$) is presented as the final indicator of the performance evaluation.

\begin{equation}
    w_{avg} = \frac{\sum_{i=1}^{5}{x_i w_i}}{\sum_{i=1}^{5}{w_i}} 
\end{equation}

\begin{equation}
    E_d = (w_{avg} - x_i)^2, \quad d = \{content, style\} 
\end{equation}

Tab. 3 and 4 show the results of the proposed performance indicators based on PSNR and SSIM, which were evaluated based on 1,452 generated results. PSNR values for content and style images are represented by $P_Content$ and $P_Style$ respectively, whereas $S_Content$ and $S_Style$ calculate the SSIM values for content and style images, respectively. $E_Content+E_Style$, which is the sum of the square errors for content and style, is used as the final metric. The preservation of content was highest when $L_h$ was not used, while not using $L_c$ or $L_s$ resulted in a loss of content and style. $L_l$ showed no significant difference in terms of content, whereas style was relatively high. Therefore, $E_PSNR$ and $E_SSIM$ are good evaluation metrics when all loss functions are used. The distribution of the results generated with the content retention performance as the style transfer performance is shown in Fig. 12. When PSNR is considered, the distribution of results with respect to $L_h$ is different from the other results. The distribution with respect to $L_c$ is located in the first half and with respect to $L_s$ in the second half. However, the distribution of $L_Total$ is relatively close to the center with a small deviation, making it the most appropriate result. In the case of SSIM, the distribution shape is similar to that of the PSNR, but several distributions show parallel movement results. The smaller the $E_SSIM$, the more central the distribution, indicating better performance. Therefore, $L_Total$ shows better performance than $L_l$, and $L_Total$ with a small difference has similar distributions in the center. Other results are considered to produce relatively poor results because they are located away from the center.




\begin{table}[htbp]
    \caption{Ablation Study Analysis (PSNR)}
        \centering   
        \begin{tabular}{cccccc}
            \hline
             & \multicolumn{5}{c}{PSNR} \\
            \cline{2-6} 
             & w.o $L_c$ & w.o $L_s$ & w.o $L_l$ & w.o $L_h$ & $L_{Total}$ \\
            \hline
            $P_{Content}$ & 8.690 & 9.833 & 9.242 & 17.350 & 9.280 \\
            $P_{Style}$ & 19.812 & 14.970 & 18.568 & 9.791 & 17.642 \\
            $E_{Content}$ & 1.745 & 0.032 & 0.591 & 53.868 & 0.534\\
            $E_{Style}$ & 9.045 & 3.368 & 3.109 & 49.301 & 0.701 \\
            \midrule
            $E_{PSNR}$ & 10.789 & 3.399 & 3.699 & 103.069 & \textbf{1.235}\\
            \hline
    \end{tabular}
\end{table}


\begin{table}[htbp]
    \caption{Ablation Study Analysis (SSIM)}
        \centering   
        \begin{tabular}{cccccc}
            \hline
             & \multicolumn{5}{c}{SSIM} \\
            \cline{2-6} 
             & w.o $L_c$ & w.o $L_s$ & w.o $L_l$ & w.o $L_h$ & $L_{Total}$ \\
            \hline
            $P_{Content}$ & 0.394 & 0.599 & 0.517 & 0.772 & 0.504 \\
            $P_{Style}$ & 0.715 & 0.466 & 0.599 & 0.338 & 0.597 \\
            $E_{Content}$ & 0.022 & 0.003 & 0.001 & 0.053 & 0.001 \\
            $E_{Style}$ & 0.025 & 0.009 & 0.002 & 0.049 & 0.001 \\
            \midrule
            $E_{PSNR}$  & 0.047 & 0.012 & 0.003 & 0.101 & \textbf{0.002} \\
            \hline
    \end{tabular}
\end{table}


\begin{figure}[htb!]
    \centering
    \includegraphics[width=14cm]{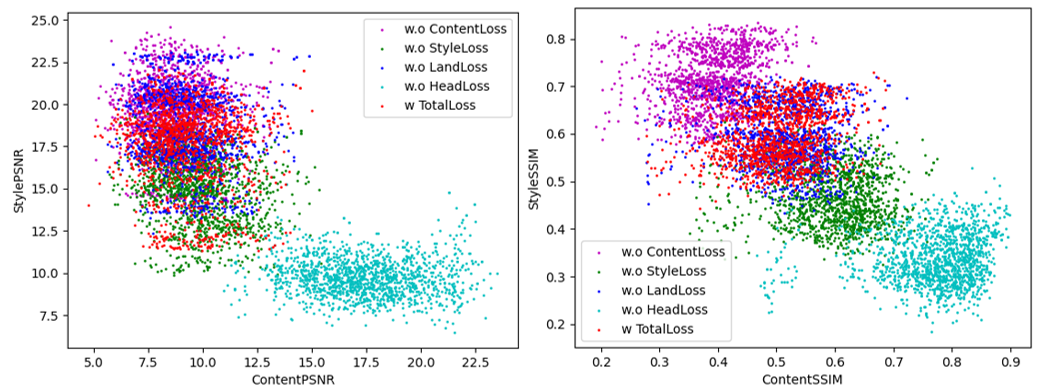}
    \caption{The results of content and style represent points for the test datasets through a two-dimensional coordinate system (top: PSNR, bottom: SSIM)}
\end{figure}




\section{Conclusions}
The objective of this study is to propose a generative adversarial network that utilizes facial feature points and loss functions to achieve arbitrary style transfers while maintaining the original face shape and transferring the Gat. To preserve the characteristics of the face, two loss functions, Land Loss and Head Loss, were defined using landmark masks to minimize the difference and speed up the learning process. Style Loss, which uses a gram matrix for content loss and style transfer, enables style transfer while preserving the character's shape. However, if the input images have large differences and the feature maps have significant discrepancies, the results are not satisfactory, and there are color differences in some instances. To overcome these limitations, it is recommended to define a loss function that considers color differences and aligns the feature map through the alignment of facial landmarks in future studies.



\bibliographystyle{unsrt}  

\begin{thebibliography}{1}

\bibitem{encykorea}
Encyclopedia of Korean Culture.
\newblock Available online: \url{http://encykorea.aks.ac.kr/Contents/Item/E0057016} (accessed on 2 May 2023).

\bibitem{si2020style}
J. Si, J. Jeong, G. Kim, and S. Kim, "Style Interconversion of Korean Portrait and ID Photo Using CycleGAN," \textit{Proc. of Korean Institute of Information Technology (KIIT)}, pages 147-149, 2020.

\bibitem{zhu2017unpaired}
J. Zhu, T. Park, P. Isola, and A. A. Efros, "Unpaired Image-To-Image Translation Using Cycle-Consistent Adversarial Networks," \textit{Proc. of the IEEE International Conf. on Computer Vision (ICCV)}, pages 2223-2232, 2017.

\bibitem{huang2017arbitrary}
X. Huang and S. Belongie, "Arbitrary Style Transfer in Real-Time With Adaptive Instance Normalization," \textit{Proc. of the IEEE International Conf. on Computer Vision (ICCV)}, pages 1501-1510, 2017.

\bibitem{huang2020parameter}
S. Huang, H. Xiong, T. Wang, Q. Wang, Z. Chen, J. Huan, and D. Dou, "Parameter-Free Style Projection for Arbitrary Style Transfer," \textit{arXiv preprint arXiv:2003.07694}, 2020.

\bibitem{zhu2020detail}
T. Zhu and S. Liu, "Detail-Preserving Arbitrary Style Transfer," \textit{Proc. of IEEE International Conf. on Multimedia and Expo (ICME)}, pages 1-6, 2020.

\bibitem{elad2017style}
M. Elad and P. Milanfar, "Style Transfer Via Texture Synthesis," \textit{IEEE Transactions on Image Processing}, vol. 26, no. 5, pages 2338-2351, 2017.

\bibitem{simonyan2015very}
K. Simonyan and A. Zisserman, "Very Deep Convolutional Networks for Large-Scale Image Recognition," \textit{Proc. of International Conf. on Learning Representations (ICLR)}, pages 1-14, 2015.

\bibitem{li2017laplacian}
S. Li, X. Xu, L. Nie, and T. Chua, "Laplacian-Steered Neural Style Transfer," \textit{Proc. of ACM international conf. on Multimedia}, pages 1716-1724, 2017.

\bibitem{chen2018face}
C. Chen, X. Tan, and K. Y. K. Wong, "Face Sketch Synthesis with Style Transfer Using Pyramid Column Feature," \textit{Proc. of IEEE Winter Conf. on Applications of Computer Vision (WACV)}, pages 485-493, 2018.

\bibitem{blakeslee2018faster}
B. Blakeslee, R. Ptucha, and A. Savakis, "FASTER ART-CNN: AN EXTREMELY FAST STYLE TRANSFER NETWORK," \textit{Proc. of IEEE Western New York Image and Signal Processing Workshop (WNYISPW)}, pages 1-5, 2018.

\bibitem{liu2020geometric}
X. Liu, X. Li, M. Cheng, and P. Hall, "Geometric style transfer," \textit{arXiv preprint arXiv:2007.05471}, 2020.

\bibitem{kaur2019photo}
P. Kaur, H. Zhang, and K. Dana, "Photo-Realistic Facial Texture Transfer," \textit{Proc. of IEEE Conf. on Applications of Computer Vision (WACV)}, pages 2097-2105, 2019.

\bibitem{yi2019apdrawinggan}
R. Yi, Y. Liu, Y. Lai, and P. Rosin, "APDrawingGAN: Generating Artistic Portrait Drawings From Face Photos With Hierarchical GANs," \textit{Proc. of the IEEE/CVF Conf. on Computer Vision and Pattern Recognition (CVPR)}, pages 10743-10752, 2019.

\bibitem{xu2019learning}
Z. Xu, M. Wilber, C. Fang, A. Hertzmann, and H. Jin, "Learning from multi-domain artistic images for arbitrary style transfer," \textit{Proc. of the ACM/Eurographics Expressive Symposium on Computational Aesthetics and Sketch Based Interfaces and Modeling and Non-Photorealistic Animation and Rendering (Expressive '19)}, pages 21-31, 2019.

\bibitem{zhang2018style}
R. Zhang, S. Tang, Y. Li, J. Guo, Y. Zhang, J. Li, and S. Yan, "Style Separation and Synthesis via Generative Adversarial Networks," \textit{Proc. of the ACM International Conf. on Multimedia}, pages 183-191, 2018.

\bibitem{horita2022slgan}
D. Horita and K. Aizawa, "SLGAN: Style- and Latent-guided Generative Adversarial Network for Desirable Makeup Transfer and Removal," \textit{Proc. Of the ACM International Conf. on Multimedia in Asia}, pages 1-8, 2022.

\bibitem{li2018beautygan}
T. Li, R. Qian, C. Dong, S. Liu, Q. Yan, W. Zhu, and L. Lin, "BeautyGAN: Instance-level Facial Makeup Transfer with Deep Generative Adversarial Network," \textit{Proc. of the ACM international conf. on Multimedia}, pages 645-653, 2018.

\bibitem{chang2018pairedcyclegan}
H. Chang, J. Lu, F. Yu, and A. Finkelstein, "PairedCycleGAN: Asymmetric Style Transfer for Applying and Removing Makeup," \textit{Proc. of the IEEE Conf. on Computer Vision and Pattern Recognition (CVPR)}, pages 40-48, 2018.

\bibitem{wu2019landmark}
R. Wu, X. Gu, X. Tao, X. Shen, Y. W. Tai, and J. Jia, "Landmark Assisted CycleGAN for Cartoon Face Generation," \textit{arXiv preprint arXiv:1907.01424}, 2019.



\bibitem{palsson2018generative}
S. Palsson, E. Agustsson, R. Timofte, and L. Van Gool, "Generative Adversarial Style Transfer Networks for Face Aging," \textit{Proc. of the IEEE Conf. on Computer Vision and Pattern Recognition Workshops (CVPRW)}, pages 2084-2092, 2018.

\bibitem{wang2019how}
Z. Wang, Z. Liu, J. Huang, S. Lian, and Y. Lin, "How Old Are You? Face Age Translation with Identity Preservation Using GANs," \textit{arXiv preprint arXiv:1909.04988}, 2019.

\bibitem{yi2020unpaired}
R. Yi, Y. J. Liu, Y. K. Lai, and P. L. Rosin, "Unpaired portrait drawing generation via asymmetric cycle mapping," \textit{Proc. of the IEEE/CVF Conf. on Computer Vision and Pattern Recognition (CVPR)}, pages 8217-8225, 2020.

\bibitem{gatys2016image}
L. A. Gatys, A. S. Ecker, and M. Bethge, "Image Style Transfer Using Convolutional Neural Networks," \textit{Proc. of the IEEE Conf. on Computer Vision and Pattern Recognition (CVPR)}, pages 2414-2423, 2016.

\bibitem{kazemi2014one}
V. Kazemi and J. Sullivan, "One millisecond face alignment with an ensemble of regression trees," \textit{Proc. of the IEEE Conf. on Computer Vision and Pattern Recognition (CVPR)}, pages 1867-1874, 2014.

\bibitem{miyato2018spectral}
T. Miyato, T. Kataoka, M. Koyama, and Y. Yoshida, "Spectral Normalization for Generative Adversarial Networks," \textit{Proc. of the International Conf. on Learning Representations (ICLR)}, pages 1-10, 2018.

\bibitem{li2016precomputed}
C. Li and M. Wand, "Precomputed real-time texture synthesis with markovian generative adversarial networks," \textit{Proc. of the European Conf. on Computer Vision (ECCV)}, pages 702-716, 2016.

\bibitem{si2021traffic}
J. Si and S. Kim, "Traffic Accident Detection in First-Person Videos based on Depth and Background Motion Estimation," \textit{Journal of Korean Institute of Information Technology (JKIIT)}, vol. 19, no. 3, pages 25-34, 2021.

\bibitem{hore2010image}
A. Hor{'e} and D. Ziou, "Image Quality Metrics: PSNR vs. SSIM," \textit{Proc. of the International Conf. on Pattern Recognition (ICPR)}, pages 2366-2369, 2010.

\end{thebibliography}

\end{document}

\end{document}